\pdfoutput=1

\documentclass{article}

\usepackage[nonatbib,final]{nips_2016}

\usepackage[T1]{fontenc}
\usepackage{tgtermes}
\usepackage[T1]{fontenc}
\usepackage{scalefnt,letltxmacro}
\LetLtxMacro{\oldtextsc}{\textsc}
\renewcommand{\textsc}[1]{\oldtextsc{\scalefont{1.10}#1}}

\usepackage{times,amsmath,amssymb,paralist,graphicx,subfig,bm,booktabs,url}
\usepackage[usenames,dvipsnames]{xcolor}
\usepackage[acronym,smallcaps,nowarn]{glossaries}
\usepackage{natbib}
\usepackage[algoruled]{algorithm2e}

\newcommand{\parhead}[1]{\textbf{#1}\,}

\newacronym{ADVI}{advi}{automatic differentiation variational inference}

\newacronym{BBVI}{bbvi}{black-box variational inference}

\newacronym{CDF}{cdf}{cumulative density function}
\newacronym{CTM}{ctm}{correlated topic model}

\newacronym[\glslongpluralkey={deep exponential families}]{DEF}{def}{deep exponential family}
\newacronym{DMIS}{dmis}{deterministic multiple importance sampling}

\newacronym{ELBO}{elbo}{evidence lower bound}

\newacronym{GNTS}{gn-ts}{gamma-normal time series model}
\newacronym{G-REP}{g-rep}{generalized reparameterization}

\newacronym{KL}{kl}{{K}ullback-{L}eibler}

\newacronym{LDA}{lda}{latent {D}irichlet allocation}

\newacronym{MF}{mf}{matrix factorization}
\newacronym{MIS}{mis}{multiple importance sampling}

\newacronym{OBBVI}{o-bbvi}{overdispersed black-box variational inference}

\newacronym{SVI}{svi}{stochastic variational inference}

\newacronym{VI}{vi}{variational inference}

\DeclareRobustCommand{\KL}[2]{\ensuremath{D_{\textrm{KL}}\left(#1\;\|\;#2\right)}}

\DeclareRobustCommand{\E}[2]{\mathbb{E}_{#1}\left[#2\right]}
\DeclareRobustCommand{\ent}[1]{\mathbb{H}\left[#1\right]}

\DeclareRobustCommand{\logit}[1]{\textrm{logit}\left(#1\right)}
\DeclareRobustCommand{\sigmoid}[1]{\textrm{sigmoid}\left(#1\right)}

\newcommand{\g}{\, | \,}

\newcommand{\Lcal}{\mathcal{L}}

\newcommand{\Tcal}{\mathcal{T}}

\newcommand{\bzero}{\mathbf{0}}
\newcommand{\bone}{\mathbf{1}}
\newcommand{\bz}{\mathbf{z}}
\newcommand{\bx}{\mathbf{x}}
\newcommand{\bg}{\mathbf{g}}

\newcommand{\bv}{\mathbf{v}}
\newcommand{\bV}{\mathbf{V}}
\newcommand{\bD}{\mathbf{D}}
\newcommand{\beps}{\bm{\epsilon}}

\newcommand{\bmu}{\bm{\mu}}
\newcommand{\brho}{\bm{\rho}}
\newcommand{\balpha}{\bm{\alpha}}
\newcommand{\bSigma}{\bm{\Sigma}}

\newcommand{\hfun}{h}
\newcommand{\ufun}{u}

\title{The Generalized Reparameterization Gradient}

\author{
  Francisco J.~R.~Ruiz\\
  University of Cambridge\\
  Columbia University\\
  \And
  Michalis K.~Titsias\\
  Athens University of\\Economics and Business\\
  \And
  David M.~Blei\\
  Columbia University\\
}

\begin{document}

\maketitle

\vspace*{-14pt}
\begin{abstract} \vspace*{-5pt}
  The reparameterization gradient has become a widely used method to obtain Monte Carlo gradients to optimize the variational objective. However, this technique does not easily apply to commonly used distributions such as beta or gamma without further approximations, and most practical applications of the reparameterization gradient fit Gaussian distributions. In this paper, we introduce \emph{the generalized reparameterization gradient}, a method that extends the reparameterization gradient to a wider class of variational distributions. Generalized reparameterizations use invertible transformations of the latent variables which lead to transformed distributions that weakly depend on the variational parameters. This results in new Monte Carlo gradients that combine reparameterization gradients and score function gradients. We demonstrate our approach on variational inference for two complex probabilistic models. The generalized reparameterization is effective: even a single sample from the variational distribution is enough to obtain a low-variance gradient.
\end{abstract}


\setlength{\textfloatsep}{7pt}


\vspace*{-10pt}
\section{Introduction}
\vspace*{-5pt}

\Gls{VI} is a technique for approximating
the posterior distribution in probabilistic models
\citep{Jordan1999,Wainwright2008}.
Given a probabilistic model $p(\bx, \bz)$ of observed variables $\bx$ and hidden variables $\bz$, the goal of \gls{VI} is to approximate the posterior
$p(\bz \g \bx)$, which is intractable to compute exactly for
many models. The idea of \gls{VI} is to posit
a family of distributions over the latent variables $q(\bz ; \bv)$
with free variational parameters $\bv$. \gls{VI} then fits those parameters
to find the member of the family that is closest in \gls{KL} divergence to the
exact posterior,
$\bv^* = \arg \min_{\bv} \textrm{KL}(q(\bz ; \bv) || p(\bz \g \bx))$.
This turns inference into optimization, and different ways of doing
\gls{VI} amount to different optimization algorithms for
solving this problem.

For a certain class of probabilistic models, those where each
conditional distribution is in an exponential family, we can easily
use coordinate ascent optimization to minimize the \gls{KL}
divergence~\citep{Ghahramani2001}. However, many important models do
not fall into this class (e.g., probabilistic neural networks or
Bayesian generalized linear models).
This is the scenario that we focus on in this paper.
Much recent research in \gls{VI} has focused on
these difficult settings, seeking effective optimization algorithms
that can be used with any model. This has enabled the application of
\gls{VI} on nonconjugate probabilistic models
\citep{Carbonetto2009,Paisley2012,Ranganath2014,Titsias2014_doubly},
deep neural networks
\citep{Neal1992,Hinton1995,Mnih2014,Kingma2014}, and probabilistic
programming \citep{Wingate2013,Kucukelbir2015,vandeMeent2016}.


One strategy for \gls{VI} in nonconjugate models is to 
obtain Monte Carlo estimates of the gradient of the variational objective
and to use stochastic optimization to fit the variational parameters.
Within this strategy, there have been two main lines of research:
\gls{BBVI}~\citep{Ranganath2014} and reparameterization
gradients~\citep{Salimans2013,Kingma2014}. Each enjoys different
advantages and limitations.

\gls{BBVI} expresses the gradient of the
variational objective as an expectation with respect to the
variational distribution using the log-derivative trick,
also called \textsc{reinforce} or score function method
\citep{Glynn1990,Williams1992}. It then takes samples from the variational
distribution to calculate noisy gradients. \gls{BBVI} is generic---it
can be used with any type of latent variables and any model. However,
the gradient estimates typically suffer from high variance, which can
lead to slow convergence. \citet{Ranganath2014} reduce the variance of
these estimates using 
Rao-Blackwellization~\citep{Casella1996} and control variates
\citep{Ross2002,Paisley2012,Gu2016}. Other researchers have proposed
further reductions, e.g., through local expectations
\citep{Titsias2015} and importance sampling \citep{Ruiz2016}.


The second approach to Monte Carlo gradients of the variational
objective is through reparameterization
\citep{Price1958,Bonnet1964,Salimans2013,Kingma2014,Rezende2014}.
This approach reparameterizes the latent variable $\bz$ in terms of a set of
auxiliary random variables whose distributions do not depend on the
variational parameters (typically, a standard normal). This facilitates
taking gradients of the
variational objective because the gradient operator can be pushed
inside the expectation, and because the resulting procedure only
requires drawing samples from simple distributions, such as standard normals.
We describe this in detail in Section~\ref{sec:background}.

Reparameterization gradients exhibit lower variance than \gls{BBVI}
gradients. They typically need only one Monte Carlo sample to estimate
a noisy gradient, which leads to fast algorithms. Further, for some
models, their variance can be bounded~\citep{Fan2015}. However,
reparameterization is not as generic
as \gls{BBVI}. It is typically used with Gaussian variational
distributions and does not easily generalize to other common
distributions, such as the gamma or beta, without using further approximations. (See \citet{Knowles2015} for an alternative approach to deal with the gamma distribution.)


We develop \textit{the \gls{G-REP}
gradient}, a new method to extend reparameterization to other
variational distributions. The main idea is to define an invertible
transformation of the latent variables such that the distribution of the
transformed variables is only weakly governed by the variational parameters. (We make this precise in Section~\ref{sec:g-rep}.) Our technique naturally combines both
\gls{BBVI} and reparameterization; it applies to a wide class of
nonconjugate models; it maintains the black-box criteria of reusing
variational families; and it avoids approximations. We empirically
show in two probabilistic models---a nonconjugate factorization
model and a deep exponential family~\citep{Ranganath2015}---that a
single Monte Carlo sample is enough to build an effective low-variance
estimate of the gradient. In terms of speed, \gls{G-REP}
outperforms \gls{BBVI}. In terms of accuracy, it outperforms
\gls{ADVI} \citep{Kucukelbir2016}, which considers Gaussian variational
distributions on a transformed space.


\vspace*{-7pt}
\section{Background}
\vspace*{-5pt}
\label{sec:background}

Consider a probabilistic model $p(\bx,\bz)$, where $\bz$ denotes the latent variables and 
$\bx$ the observations. We assume that the posterior distribution $p(\bz\g\bx)$ is analytically intractable
and we wish to apply \gls{VI}. We introduce a
tractable distribution $q(\bz;\bv)$ to approximate $p(\bz\g\bx)$ and minimize the \gls{KL} divergence $\KL{q(\bz;\bv)}{p(\bz\g \bx)}$ with respect to the variational parameters $\bv$. This minimization is equivalently expressed as the maximization of the so-called \gls{ELBO} \citep{Jordan1999},
\begin{equation}\label{eq:elbo}
  \Lcal(\bv) = \E{q(\bz;\bv)}{\log p(\bx,\bz)-\log q(\bz;\bv)} = \E{q(\bz;\bv)}{f(\bz)} + \ent{q(\bz;\bv)}.
\end{equation}
We denote
\begin{equation}
  f(\bz)\triangleq\log p(\bx,\bz)
\end{equation}
to be the model log-joint density and $\ent{q(\bz;\bv)}$ to be the entropy of the variational distribution.
When the expectation $\E{q(\bz;\bv)}{f(\bz)}$ is analytically tractable, the maximization 
of the \gls{ELBO} can be carried out using standard optimization methods. Otherwise, when it is intractable, other techniques are needed. Recent approaches rely on stochastic optimization to construct Monte Carlo estimates of the gradient with respect to the variational parameters. Below, we review the two main methods for building such Monte Carlo estimates: the score function method and the reparameterization trick. 

\parhead{Score function method.} A general way to obtain unbiased stochastic gradients is to use the 
score function method, also called log-derivative trick or \textsc{reinforce} \citep{Williams1992,Glynn1990}, which has been recently applied to \gls{VI} \citep{Paisley2012,Ranganath2014,Mnih2014}. It is based on writing the gradient of the \gls{ELBO} with respect to $\bv$ as
\begin{equation}\label{eq:reinforce}
  \nabla_{\bv}\Lcal = \E{q(\bz;\bv)}{f(\bz) \nabla_{\bv}\log q(\bz;\bv)} + \nabla_{\bv}\ent{q(\bz;\bv)},
\end{equation}
and then building Monte Carlo estimates by approximating the expectation with samples from $q(\bz;\bv)$. 
The resulting estimator suffers from high variance, making it necessary to apply variance reduction methods such as control variates \citep{Ross2002} or Rao-Blackwellization \citep{Casella1996}. Such variance reduction techniques have been used in \gls{BBVI} \citep{Ranganath2014}.

\parhead{Reparameterization.} The reparameterization trick \citep{Salimans2013,Kingma2014} expresses the latent variables $\bz$ as an invertible function of another set of variables $\beps$, i.e., $\bz=\Tcal(\beps;\bv)$, such that the distribution of the new random variables $q_{\beps}(\beps)$ does not depend on the variational parameters $\bv$. Under these assumptions, expectations with respect to $q(\bz;\bv)$ can be expressed as $\E{q(\bz;\bv)}{f(\bz)}=\E{q_{\beps}(\beps)}{f\left(\Tcal(\beps;\bv)\right)}$, and the gradient with respect to $\bv$ can be pushed into the expectation, yielding
\begin{equation}\label{eq:reparam}
  \nabla_{\bv}\Lcal = \E{q_{\beps}(\beps)}{\nabla_{\bz}f(\bz)\big|_{\bz=\Tcal(\beps;\bv)} \nabla_{\bv} \Tcal(\beps;\bv)} + \nabla_{\bv}\ent{q(\bz;\bv)}.
\end{equation}
The assumption here is that the log-joint $f(\bz)$ is differentiable with respect to the latent variables. The gradient $\nabla_{\bz}f(\bz)$ depends on the model, but it can be computed using automatic differentiation tools \citep{Baydin2015}. Monte Carlo estimates of the reparameterization gradient typically present much lower variance than those based on Eq.~\ref{eq:reinforce}. In practice, a single sample from $q_{\beps}(\beps)$ is enough to estimate a low-variance gradient.\footnote{In the literature, there is no formal proof that reparameterization has lower variance than the score function estimator, except for some simple models \citep{Fan2015}. \citet{Titsias2014_doubly} provide some intuitions, and \citet{Rezende2014} show some benefits of reparameterization in the Gaussian case.}

The reparameterization trick is thus a powerful technique to reduce the variance of the estimator, but it requires a transformation $\beps=\Tcal^{-1}(\bz;\bv)$ such that $q_{\beps}(\beps)$ does not depend on the variational parameters $\bv$. For instance, if the variational distribution is Gaussian with mean $\bmu$ and covariance $\bSigma$, a straightforward transformation consists of standardizing the random variable $\bz$, i.e.,
\begin{equation}\label{eq:standardiz_gaussian}
  \beps=\Tcal^{-1}(\bz;\bmu,\bSigma)=\bSigma^{-\frac{1}{2}}(\bz-\bmu).
\end{equation}
This transformation ensures that the (Gaussian) distribution $q_{\beps}(\beps)$ does not depend on $\bmu$ or $\bSigma$. For a general variational distribution $q(\bz;\bv)$, \citet{Kingma2014} discuss three families of transformations: inverse \gls{CDF}, location-scale, and composition. However, these transformations may not apply in certain cases.\footnote{The inverse \gls{CDF} approach sets $\Tcal^{-1}(\bz;\bv)$ to the \gls{CDF}. This leads to a uniform distribution over $\beps$ on the unit interval, but it is not practical because the inverse \gls{CDF}, $\Tcal(\beps;\bv)$, does not have analytical solution in general. We develop an approach that does not require computation of \glspl{CDF} or their derivatives.}
Notably, none of them apply to the gamma\footnote{Composition is only available when it is possible to express the gamma as a sum of exponentials, i.e., its shape parameter is an integer, which is not generally the case in \gls{VI}.} and the beta distributions, although these distributions are often used in \gls{VI}.



Next, we show how to relax the constraint that the transformed density $q_{\beps}(\beps)$ must not depend on the variational parameters $\bv$. We follow a standardization procedure similar to the Gaussian case in Eq.~\ref{eq:standardiz_gaussian}, but we allow the distribution of the standardized variable $\beps$ to depend (at least weakly) on $\bv$.


\vspace*{-5pt}
\section{The Generalized Reparameterization Gradient}
\vspace*{-5pt}
\label{sec:g-rep}

We now generalize the reparameterization idea to distributions that, like the gamma or the beta, do not admit the standard reparameterization trick. We assume that we can efficiently sample from the variational distribution $q(\bz;\bv)$, and that $q(\bz;\bv)$ is differentiable with respect to $\bz$ and $\bv$.
We introduce a random variable $\beps$ defined by an invertible transformation 
\begin{equation}
\beps = \Tcal^{-1}(\bz;\bv), \qquad \textrm{and} \qquad \bz=\Tcal(\beps;\bv),
\end{equation} 
where we can think of $\beps = \Tcal^{-1}(\bz;\bv)$ as a \emph{standardization procedure} that 
attempts to make the distribution of $\beps$ weakly dependent on the variational parameters $\bv$. ``Weakly'' means that at least its first moment does not depend on $\bv$. For instance, if $\beps$ is defined to have zero mean, then its first moment has become independent of $\bv$. However, we \emph{do not} assume that the resulting distribution of $\beps$ is completely independent of the variational parameters $\bv$, and therefore we write it as $q_{\beps}(\beps;\bv)$. We use the distribution $q_{\beps}(\beps;\bv)$ in the derivation of \gls{G-REP}, but we write the final gradient as an expectation with respect to the original variational distribution $q(\bz;\bv)$, from which we can sample. 

More in detail, by the standard change-of-variable technique, the transformed density is
\begin{equation}\label{eq:q_eps}
  q_{\beps}(\beps;\bv)=q\left(\Tcal(\beps;\bv);\bv\right)J(\beps,\bv),\quad {\rm where} \quad
  J(\beps,\bv)\triangleq\left|\det \nabla_{\beps}\Tcal(\beps;\bv)\right|,
\end{equation}
is a short-hand for the absolute value of the determinant of the Jacobian. We first use the transformation to rewrite the gradient of $\E{q(\bz;\bv)}{f(\bz)}$ in \eqref{eq:elbo} as
\begin{equation}
  \begin{split}
      \nabla_{\bv} \E{q(\bz;\bv)}{f(\bz)} = \nabla_{\bv}\E{q_{\beps}(\beps;\bv)}{f\left(\Tcal(\beps;\bv)\right)} = \nabla_{\bv}\int q_{\beps}(\beps;\bv)f\left(\Tcal(\beps;\bv)\right) d\beps.
  \end{split}
\end{equation}
We now express the gradient as the sum of two terms, which we name $\bg^{\textrm{rep}}$ and $\bg^{\textrm{corr}}$ for reasons that we will explain below. We apply the log-derivative trick and the product rule for derivatives, yielding
\begin{equation}\label{eq:grad_as_sum_of_two_terms}
  \begin{split}
      \nabla_{\bv} \E{q(\bz;\bv)}{f(\bz)} = \underbrace{\int q_{\beps}(\beps;\bv)\nabla_{\bv}f\left(\Tcal(\beps;\bv)\right)d\beps}_{\bg^{\textrm{rep}}} +
      \underbrace{\int q_{\beps}(\beps;\bv)f\left(\Tcal(\beps;\bv)\right)\nabla_{\bv} \log q_{\beps}(\beps;\bv) d\beps}_{\bg^{\textrm{corr}}},
  \end{split}
\end{equation}
We rewrite Eq.~\ref{eq:grad_as_sum_of_two_terms} as an expression that involves expectations with respect to the original variational distribution $q(\bz;\bv)$ only. For that, we define the following two auxiliary functions that depend on the transformation $\Tcal(\beps;\bv)$:
\begin{equation}\label{eq:hfun_ufun}
    \hfun(\beps;\bv)\triangleq \nabla_{\bv} \Tcal(\beps;\bv),
    \qquad \textrm{and} \qquad
    \ufun(\beps;\bv)\triangleq \nabla_{\bv} \log J(\beps,\bv).
\end{equation}
After some algebra (see the Supplement for details), we obtain
\begin{equation}\label{eq:grep_gcorr}
  \begin{split}
      & \bg^{\textrm{rep}} = \E{q(\bz;\bv)}{\nabla_{\bz}f(\bz) \hfun\left( \Tcal^{-1}(\bz;\bv);\bv \right)},\\
      & \bg^{\textrm{corr}} = \E{q(\bz;\bv)}{f(\bz) \left(\nabla_{\bz}\log q(\bz;\bv) \hfun\left(\Tcal^{-1}(\bz;\bv);\bv\right) + \nabla_{\bv}\log q(\bz;\bv)+ \ufun\left(\Tcal^{-1}(\bz;\bv);\bv \right) \right)}.
  \end{split}
\end{equation}
Thus, we can finally write the full gradient of the \gls{ELBO} as
\begin{equation}\label{eq:corrected_reparam}
  \nabla_{\bv}\Lcal = \bg^{\textrm{rep}} + \bg^{\textrm{corr}} + \nabla_{\bv}\ent{q(\bz;\bv)},
\end{equation}

\vspace*{-5pt}
\parhead{Interpretation of the generalized reparameterization gradient.}
The term $\bg^{\textrm{rep}}$ is easily recognizable as the standard reparameterization gradient, and hence the label ``rep.'' Indeed, if the distribution $q_{\beps}(\beps;\bv)$ does not depend on the variational parameters $\bv$, then the term $\nabla_{\bv} \log q_{\beps}(\beps;\bv)$ in Eq.~\ref{eq:grad_as_sum_of_two_terms} vanishes, making $\bg^{\textrm{corr}}=\bzero$. Thus, we may interpret $\bg^{\textrm{corr}}$ as a ``correction'' term that appears when the transformed density depends on the variational parameters.

Furthermore, we can recover the score function gradient in Eq.~\ref{eq:reinforce} by choosing the identity transformation, $\bz = \Tcal(\beps;\bv) = \beps$. In such case, the auxiliary functions in Eq.~\ref{eq:hfun_ufun} become zero because the transformation does not depend on $\bv$, i.e., $\hfun(\beps;\bv)=\bzero$ and $\ufun(\beps;\bv)=\bzero$. This implies that $\bg^{\textrm{rep}}=\bzero$ and $\bg^{\textrm{corr}}=\E{q(\bz;\bv)}{f(\bz)\nabla_{\bv}\log q(\bz;\bv)}$.


Alternatively, we can interpret the \gls{G-REP} gradient as a control variate of the score function gradient. For that, we rearrange Eqs.~\ref{eq:grad_as_sum_of_two_terms} and \ref{eq:grep_gcorr} to express the gradient as
\begin{align}
  \nabla_{\bv} \E{q(\bz;\bv)}{f(\bz)} = & \; \E{q(\bz;\bv)}{f(\bz) \nabla_{\bv}\log q(\bz;\bv)} \nonumber \\
  & + \bg^{\textrm{rep}} + \E{q(\bz;\bv)}{f(\bz) \left(\nabla_{\bz}\log q(\bz;\bv) \hfun\left(\Tcal^{-1}(\bz;\bv);\bv\right) + \ufun\left(\Tcal^{-1}(\bz;\bv);\bv \right) \right)},\nonumber
\end{align}
where the second line is the control variate, which involves the reparameterization gradient.

\parhead{Transformations.}
Eqs.~\ref{eq:grad_as_sum_of_two_terms} and \ref{eq:grep_gcorr} are valid for any transformation $\Tcal(\beps;\bv)$. However, we may expect some transformations to perform better than others, in terms of the variance of the resulting estimator. It seems sensible to search for transformations that make $\bg^{\textrm{corr}}$ small, as the reparameterization gradient $\bg^{\textrm{rep}}$ is known to present low variance in practice under standard smoothness conditions of the log-joint \citep{Fan2015}.\footnote{Techniques such as Rao-Blackwellization could additionally be applied to reduce the variance of $\bg^{\textrm{corr}}$. We do \emph{not} apply any such technique in this paper.} Transformations that make $\bg^{\textrm{corr}}$ small are such that $\beps = \Tcal^{-1}(\bz;\bv)$ becomes weakly dependent on the variational parameters $\bv$.
In the standard reparameterization of Gaussian random variables, the transformation takes the form in \eqref{eq:standardiz_gaussian}, and thus $\beps$ is a standardized version of $\bz$. We mimic this standardization idea for other distributions as well. In particular, for exponential family distributions, we use transformations of the form (sufficient statistic $-$ expected sufficient statistic)$/$(scale factor). We present several examples in the next section.

\vspace*{-5pt}
\subsection{Examples}
\vspace*{-5pt}
For concreteness, we show here some examples of the equations above for well-known probability distributions. In particular, we choose the gamma, log-normal, and beta distributions.

\parhead{Gamma distribution.}
Let $q(z;\alpha,\beta)$ be a gamma distribution with shape $\alpha$ and rate $\beta$. We use a transformation based on standardization of the sufficient statistic $\log(z)$, i.e.,
\begin{equation}
  \epsilon=\Tcal^{-1}(z;\alpha,\beta)=\frac{\log(z)-\psi(\alpha)+\log(\beta)}{\sqrt{\psi_1(\alpha)}},\nonumber
\end{equation}
where $\psi(\cdot)$ denotes the digamma function, and $\psi_k(\cdot)$ is its $k$-th derivative. This ensures that $\epsilon$ has zero mean and unit variance, and thus its two first moments do not depend on the variational parameters $\alpha$ and $\beta$. We now compute the auxiliary functions in Eq.~\ref{eq:hfun_ufun} for the components of the gradient with respect to $\alpha$ and $\beta$, which take the form
\begin{align}
    & \hfun_{\alpha}(\epsilon;\alpha,\beta) = \Tcal(\epsilon;\alpha,\beta) \left( \frac{\epsilon\psi_2(\alpha)}{2\sqrt{\psi_1(\alpha)}}+\psi_1(\alpha)\right),
    & & \hfun_{\beta}(\epsilon;\alpha,\beta) = -\frac{\Tcal(\epsilon;\alpha,\beta)}{\beta}, \nonumber \\
    & \ufun_{\alpha}(\epsilon;\alpha,\beta) = \left( \frac{\epsilon\psi_2(\alpha)}{2\sqrt{\psi_1(\alpha)}}+\psi_1(\alpha)\right)+\frac{\psi_2(\alpha)}{2\psi_1(\alpha)},
    & & \ufun_{\beta}(\epsilon;\alpha,\beta) = -\frac{1}{\beta}.\nonumber
\end{align}
The terms $\bg^{\textrm{rep}}$ and $\bg^{\textrm{corr}}$ are obtained after substituting these results in Eq.~\ref{eq:grep_gcorr}. We provide the final expressions in the Supplement. We remark here that the component of $\bg^{\textrm{corr}}$ corresponding to the derivative with respect to the rate equals zero, i.e., $\bg^{\textrm{corr}}_{\beta}=0$, meaning that the distribution of $\epsilon$ does not depend on the parameter $\beta$. Indeed, we can compute this distribution following Eq.~\ref{eq:q_eps} as
\begin{equation}
  q_{\epsilon}(\epsilon;\alpha,\beta)= \frac{e^{\alpha\psi(\alpha)}\sqrt{\psi_1(\alpha)}}{\Gamma(\alpha)} \exp\left(\epsilon\alpha\sqrt{\psi_1(\alpha)}-\exp\left( \epsilon\sqrt{\psi_1(\alpha)}+\psi(\alpha) \right)\right),\nonumber
\end{equation}
where we can verify that it does not depend on $\beta$.

\parhead{Log-normal distribution.}
For a log-normal distribution with location $\mu$ and scale $\sigma$, we can standardize the sufficient statistic $\log(z)$ as
\begin{equation}
  \epsilon=\Tcal^{-1}(z;\mu,\sigma)=\frac{\log(z)-\mu}{\sigma}.\nonumber
\end{equation}
This leads to a standard normal distribution on $\epsilon$, which does not depend on the variational parameters, and thus $\bg^{\textrm{corr}}=\bzero$. The auxiliary function $\hfun(\epsilon;\mu,\sigma)$, which is needed for $\bg^{\textrm{rep}}$, takes the form
\begin{align}
    & \hfun_{\mu}(\epsilon;\mu,\sigma) = \Tcal(\epsilon;\mu,\sigma),
    & \hfun_{\sigma}(\epsilon;\mu,\sigma) = \epsilon\Tcal(\epsilon;\mu,\sigma). \nonumber
\end{align}
Thus, the reparameterization gradient is given in this case by
\begin{align}
    & \bg^{\textrm{rep}}_{\mu} = \E{q(z;\mu,\sigma)}{z\nabla_{z}f(\bz)},
    & \bg^{\textrm{rep}}_{\sigma} = \E{q(z;\mu,\sigma)}{z\Tcal^{-1}(z;\mu,\sigma)\nabla_{z}f(\bz)}. \nonumber
\end{align}
This corresponds to \gls{ADVI} \citep{Kucukelbir2016} with a logarithmic transformation over a positive random variable, since the variational distribution over the transformed variable is Gaussian. For a general variational distribution, we recover \gls{ADVI} if the transformation makes $\epsilon$ Gaussian.

\parhead{Beta distribution.}
For a random variable $z\sim\textrm{Beta}(\alpha,\beta)$, we could rewrite $z=z_1^\prime/(z_1^\prime+z_2^\prime)$ for $z_1^\prime\sim\textrm{Gamma}(\alpha,1)$ and $z_2^\prime\sim\textrm{Gamma}(\beta,1)$, and apply the gamma reparameterization for $z_1^\prime$ and $z_2^\prime$. Instead, in the spirit of applying standardization directly over $z$, we define a transformation to standardize the logit function, $\logit{z}\triangleq\log(z/(1-z))$ (sum of sufficient statistics of the beta),
\begin{equation}\nonumber
  \epsilon=\Tcal^{-1}(z;\alpha,\beta)=\frac{\logit{z}-\psi(\alpha)+\psi(\beta)}{\sigma(\alpha,\beta)}.
\end{equation}
This ensures that $\epsilon$ has zero mean. We can set the denominator to the standard deviation of $\logit{z}$. However, for larger-scaled models we found better performance with a denominator $\sigma(\alpha,\beta)$ that makes $\bg^{\textrm{corr}}=\bzero$ for the currently drawn sample $z$ (see the Supplement for details), even though the variance of the transformed variable $\epsilon$ is not one in such case.\footnote{Note that this introduces some bias since we are ignoring the dependence of $\sigma(\alpha,\beta)$ on $z$.} The reason is that $\bg^{\textrm{corr}}$ suffers from high variance in the same way as the score function estimator does.



\vspace*{-5pt}
\subsection{Algorithm}
\vspace*{-5pt}
We now present our full algorithm for \gls{G-REP}. It requires the specification of the variational family and the transformation $\Tcal(\beps;\bv)$. Given these, the full procedure is summarized in Algorithm~\ref{alg:g-rep}.
We use the adaptive step-size sequence proposed by \citet{Kucukelbir2016}, which combines \textsc{rmsprop} \citep{Tieleman2012} and Adagrad \citep{Duchi2011}. Let $g_k^{(i)}$ be the $k$-th component of the gradient at the $i$-th iteration, and $\rho_k^{(i)}$ the step-size for that component. We set
\begin{equation}\label{eq:step_schedule}
  \rho_k^{(i)} = \eta \times i^{-0.5+\kappa}\times \left ( \tau + \sqrt{s_k^{(i)}} \right)^{-1},\qquad \textrm{with} \qquad s_k^{(i)} = \gamma (g_k^{(i)})^2 + (1-\gamma)s_k^{(i-1)},
\end{equation}
where we set $\kappa=10^{-16}$, $\tau=1$, $\gamma=0.1$, and we explore several values of $\eta$. Thus, we update the variational parameters as $\bv^{(i+1)} = \bv^{(i)} + \brho^{(i)}\circ\nabla_{\bv}\Lcal$, where `$\circ$' is the element-wise product.

\begin{algorithm}[t]
    \DontPrintSemicolon
    \SetAlgoLined
    \SetKwInOut{KwInput}{input}
    \SetKwInOut{KwOutput}{output}
    \KwInput{data $\bx$, probabilistic model $p(\bx,\bz)$, variational family $q(\bz;\bv)$, transformation $\bz=\Tcal(\beps;\bv)$}
    \KwOutput{variational parameters $\bv$}
    Initialize $\bv$\;
    \Repeat{convergence}{
      Draw a single sample $\bz\sim q(\bz;\bv)$\;
      Compute the auxiliary functions $\hfun\left(\Tcal^{-1}(\bz;\bv);\bv\right)$ and $\ufun\left(\Tcal^{-1}(\bz;\bv);\bv\right)$ (Eq.~\ref{eq:hfun_ufun})\;
      Estimate $\bg^{\textrm{rep}}$ and $\bg^{\textrm{corr}}$ (Eq.~\ref{eq:grep_gcorr}, estimate the expectation with one sample)\;
      Compute (analytic) or estimate (Monte Carlo) the gradient of the entropy, $\nabla_{\bv}\ent{q(\bz;\bv)}$\;
      Compute the noisy gradient $\nabla_{\bv}\Lcal$ (Eq.~\ref{eq:corrected_reparam})\;
      Set the step-size $\brho^{(i)}$ (Eq.~\ref{eq:step_schedule}) and take a gradient step for $\bv$\; 
    }
    \caption{Generalized reparameterization gradient algorithm\label{alg:g-rep}}
\end{algorithm}

\vspace*{-5pt}
\subsection{Related work}
\vspace*{-6pt}
A closely related \gls{VI} method is \gls{ADVI}, which also relies on reparameterization and has been incorporated into Stan \citep{Kucukelbir2015,Kucukelbir2016}. \gls{ADVI} applies a transformation to the random variables such that their support is on the reals and then uses a Gaussian variational posterior on the transformed space. For instance, random variables that are constrained to be positive are first transformed through a logarithmic function and then a Gaussian variational approximating distribution is placed on the unconstrained space. Thus, \gls{ADVI} struggles to approximate probability densities with singularities, which are useful in models where sparsity is appropriate. In contrast, the \gls{G-REP} method allows to estimate the gradient for a wider class of variational distributions, including gamma and beta distributions, which are more appropriate to encode sparsity constraints.


\citet{Schulman2015} also write the gradient in the form given in Eq.~\ref{eq:corrected_reparam} to automatically estimate the gradient through a backpropagation algorithm in the context of stochastic computation graphs. However, they do not provide additional insight into this equation, do not apply it to general \gls{VI}, do not discuss transformations for any distributions, and do not report experiments. Thus, our paper complements \citet{Schulman2015} and provides an off-the-shelf tool for general \gls{VI}.


\vspace*{-6pt}
\section{Experiments}
\vspace*{-5pt}
We apply \gls{G-REP} to perform mean-field \gls{VI} on two nonconjugate probabilistic models: the sparse gamma \gls{DEF} and a beta-gamma \gls{MF} model.
The sparse gamma \gls{DEF} \citep{Ranganath2015} is a probabilistic model with several layers of latent locations and latent weights, mimicking the architecture of a deep neural network. The weights of the model are denoted by $w_{k k^\prime}^{(\ell)}$, where $k$ and $k^\prime$ run over latent components, and $\ell$ indexes the layer. The latent locations are $z_{nk}^{(\ell)}$, where $n$ denotes the observation. We consider Poisson-distributed observations $x_{nd}$ for each dimension $d$. Thus, the model is specified as
\vspace*{-2pt}
\begin{align}
    & z_{nk}^{(\ell)} \sim \textrm{Gamma}\left(\alpha_z, \frac{\alpha_z}{\sum_{k^\prime} z_{nk^\prime}^{(\ell+1)} w_{k^\prime k}^{(\ell)}}\right),\qquad
    x_{nd} \sim \textrm{Poisson}\left(\sum_{k^\prime} z_{nk^\prime}^{(1)} w_{k^\prime d}^{(0)}\right).\nonumber
\end{align} \vspace*{-2pt}%
We place gamma priors over the weights $w_{k k^\prime}^{\ell}$ with rate $0.3$ and shape $0.1$, and a gamma prior with rate $0.1$ and shape $0.1$ over the top-layer latent variables $z_{nk}^{(L)}$. We set the hyperparameter $\alpha_z=0.1$, and we use $L=3$ layers with $100$, $40$, and $15$ latent factors.

The second model is a beta-gamma \gls{MF} model with weights $w_{kd}$ and latent locations $z_{nk}$. We use this model to describe binary observations $x_{nd}$, which are modeled as
\vspace*{-2pt}
\begin{equation}\nonumber
    x_{nd} \sim \textrm{Bernoulli}\left(\sigmoid{\sum_{k} \logit{z_{nk}} w_{kd}}\right),
\end{equation} \vspace*{-2pt}%
where $\logit{z}=\log(z/(1-z))$ and $\sigmoid{\cdot}$ is the inverse logit function. We place a gamma prior with shape $0.1$ and rate $0.3$ over the weights $w_{kd}$, a uniform prior over the variables $z_{nk}$, and we use $K=100$ latent components.

\parhead{Datasets.}
We apply the sparse gamma \gls{DEF} on two different databases: (i) the Olivetti database at AT\&T,\footnote{\url{http://www.cl.cam.ac.uk/research/dtg/attarchive/facedatabase.html}} which consists of 400 (320 for training and 80 for test) $64\times 64$ images of human faces in a 8 bit scale ($0-255$); and (ii) the collection of papers at the Neural Information Processing Systems (\textsc{nips}) 2011 conference, which consists of $305$ documents and a vocabulary of $5715$ effective words in a bag-of-words format ($25\%$ of words from all documents are set aside to form the test set).

We apply the beta-gamma \gls{MF} on: (i) the binarized \textsc{mnist} data,\footnote{\url{http://yann.lecun.com/exdb/mnist}} which consists of $28\times 28$ images of hand-written digits (we use $5000$ training and $2000$ test images); and (ii) the Omniglot dataset \citep{Lake2015}, which consists of $105\times 105$ images of hand-written characters from different alphabets (we select 10 alphabets, with $4425$ training images, $1475$ test images, and $295$ characters).

\vspace*{-1pt}
\parhead{Evaluation.}
We apply mean-field \gls{VI} and we compare \gls{G-REP} with \gls{BBVI} \citep{Ranganath2014} and \gls{ADVI} \citep{Kucukelbir2016}. We do not apply \gls{BBVI} on the Omniglot dataset due to its computational complexity.
At each iteration, we evaluate the \gls{ELBO} using one sample from the variational distribution, except for \gls{ADVI}, for which we use $20$ samples (for the Omniglot dataset, we only use one sample). We run each algorithm with a fixed computational budget of CPU time. After that time, we also evaluate the predictive log-likelihood on the test set, averaging over 100 posterior samples. For the \textsc{nips} data, we also compute the test perplexity (with one posterior sample) every 10 iterations, given by
\vspace*{-2pt}
\begin{equation}\nonumber
  \exp\left(\frac{-\sum_{\rm docs}\sum_{w\in \textrm{doc}(d)} \log p(w\g \#\textrm{held out in doc}(d))}{\# \textrm{held out words}}\right).
\end{equation}

\vspace*{-7pt}
\parhead{Experimental setup.}
To estimate the gradient, we use $30$ Monte Carlo samples for \gls{BBVI}, and only $1$ for \gls{ADVI} and \acrshort{G-REP}. For \gls{BBVI}, we use Rao-Blackwellization and control variates (we use a separate set of $30$ samples to estimate the control variates). For \gls{BBVI} and \acrshort{G-REP}, we use beta and gamma variational distributions, whereas \gls{ADVI} uses Gaussian distributions on the transformed space, which correspond to log-normal or logit-normal distributions on the original space. Thus, only \acrshort{G-REP} and \gls{BBVI} optimize the same variational family. We parameterize the gamma distribution in terms of its shape and mean, and the beta in terms of its shape parameters $\alpha$ and $\beta$. To avoid constrained optimization, we apply the transformation $v^\prime=\log(\exp(v)-1)$ to the variational parameters that are constrained to be positive and take stochastic gradient steps with respect to $v^\prime$. We use the analytic gradient of the entropy terms. We implement \gls{ADVI} as described by \citet{Kucukelbir2016}.

We use the step-size schedule in Eq.~\ref{eq:step_schedule}, and we explore the parameter $\eta\in\{0.1,0.5,1,5\}$. For each algorithm and each dataset, we report the results based on the value of $\eta$ for which the best \gls{ELBO} was achieved. We report the values of $\eta$ in Table~\ref{tab:eta_and_runtime} (left).


\begin{table}[t]
  \small
  \centering
  \begin{tabular}{cccc}\toprule
     \textbf{Dataset} & \acrshort{G-REP} & \gls{BBVI} & \gls{ADVI} \\ \hline
     Olivetti & $5$ & $1$ & $0.1$ \\ 
     \textsc{nips} & $0.5$ & $5$ & $1$ \\ 
     \textsc{mnist} & $5$ & $5$ & $0.1$ \\ 
     Omniglot & $5$ & $-$ & $0.1$ \\ \bottomrule
  \end{tabular}
  \hspace*{30pt}
  \begin{tabular}{cccc}\toprule
     \textbf{Dataset} & \acrshort{G-REP} & \gls{BBVI} & \gls{ADVI} \\ \hline
     Olivetti     & $0.46$  & $12.90$ & $0.17$ \\ 
     \textsc{nips}  & $0.83$  & $20.95$ & $0.25$ \\ 
     \textsc{mnist} & $1.09$  & $25.99$ & $0.34$ \\ 
     Omniglot     & $5.50$  & $-$   & $4.10$ \\ \bottomrule
  \end{tabular}
  \caption{\label{tab:eta_and_runtime}(Left) Step-size constant $\eta$, reported for completeness. (Right) Average time per iteration in seconds. \acrshort{G-REP} is $1$-$4$ times slower than \gls{ADVI} but above one order of magnitude faster than \gls{BBVI}.}
  \vspace*{-9pt}
\end{table}

\begin{figure}[t]
  \centering
  \subfloat[\gls{ELBO} (Olivetti dataset).\label{fig:elbo_faces}]{\includegraphics[width=0.34\textwidth]{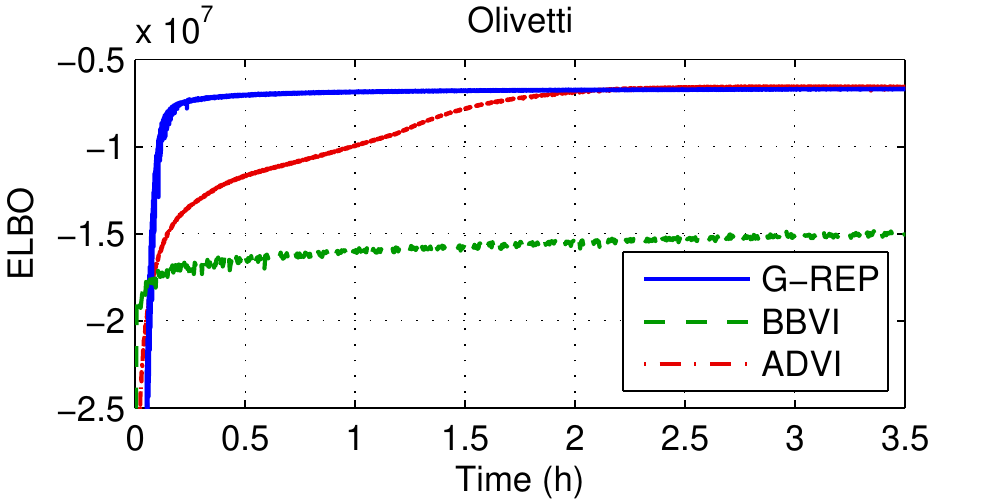}}\hspace*{-11pt}
  \subfloat[\gls{ELBO} (\textsc{mnist} dataset).\label{fig:elbo_bmnist}]{\includegraphics[width=0.34\textwidth]{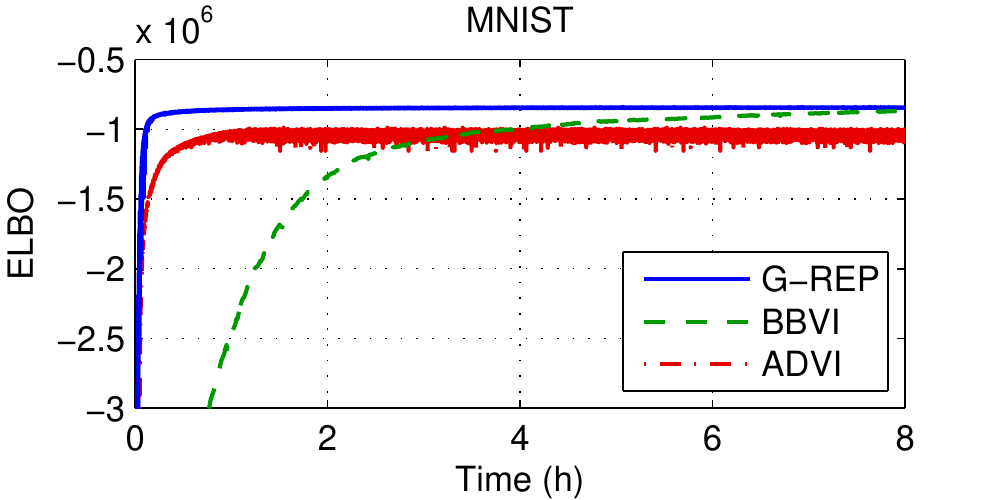}}\hspace*{-15pt}
  \subfloat[\gls{ELBO} (Omniglot dataset).\label{fig:elbo_omniglot}]{\includegraphics[width=0.34\textwidth]{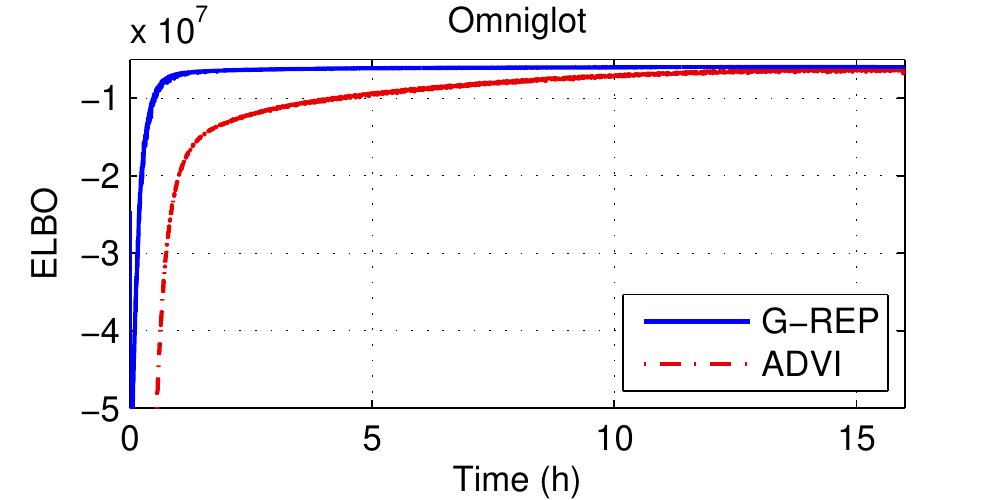}}
  \vspace*{-4pt}
  \caption{\label{fig:elbo}Comparison between \acrshort{G-REP}, \gls{BBVI}, and \gls{ADVI} in terms of the variational objective function.}
  \vspace*{-12pt}
\end{figure}

\vspace*{-1pt}
\parhead{Results.}
We show in Figure~\ref{fig:elbo} the evolution of the \gls{ELBO} as a function of the running time for three of the considered datasets. \gls{BBVI} converges slower than the rest of the methods, since each iteration involves drawing multiple samples and evaluating the log-joint for each of them. \gls{ADVI} and \acrshort{G-REP} achieve similar bounds, except for the \textsc{mnist} dataset, for which \acrshort{G-REP} provides a variational approximation that is closer to the posterior, since the \gls{ELBO} is higher. This is because a variational family with sparse gamma and beta distributions provides a better fit to the data than the variational family to which \gls{ADVI} is limited (log-normal and logit-normal). \gls{ADVI} seems to converge slower; however, we \emph{do not} claim that \gls{ADVI} converges slower than \acrshort{G-REP} in general. Instead, the difference may be due to the different step-sizes schedules that we found to be optimal (see Table~\ref{tab:eta_and_runtime}). We also report in Table~\ref{tab:eta_and_runtime} (right) the average time per iteration\footnote{%
  On the full \textsc{mnist} with $50,000$ training images, \acrshort{G-REP} (\acrshort{ADVI}) took $8.08$ ($2.04$) seconds per iteration.
} for each method: \gls{BBVI} is the slowest method, and \gls{ADVI} is the fastest because it involves simulation of Gaussian random variables only.

\begin{figure}[t]
  \centering
  \subfloat[Perplexity (\textsc{nips} dataset).\label{fig:pred_nips}]{\includegraphics[width=0.34\textwidth]{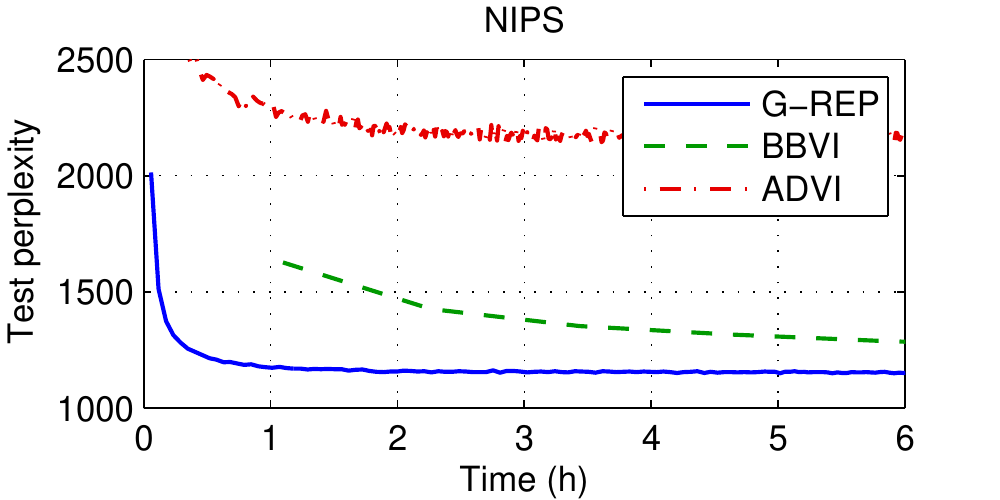}}
  \raisebox{27pt}{
    \subfloat[Average test log-likelihood per entry $x_{nd}$.\label{tab:pred_llh}]{%
      \small
      \setlength{\tabcolsep}{2.5pt}
      \begin{tabular}{cccc}\toprule
         \textbf{Dataset} & \acrshort{G-REP} & \gls{BBVI} & \gls{ADVI} \\ \hline
         Olivetti & $\mathbf{-4.48\pm 0.01}$ & $-9.74\pm 0.08$ & $-4.63\pm 0.01$ \\ 
         \textsc{mnist} & $-0.0932\pm 0.0004$ & $\mathbf{-0.0888\pm 0.0004}$ & $-0.189\pm 0.009$ \\ 
         Omniglot & $\mathbf{-0.0472\pm 0.0001}$ & $-$ & $-0.0823\pm 0.0009$ \\ \bottomrule
      \end{tabular}
    }
  }
  \vspace*{-4pt}
  \caption{\label{fig:pred_perm}Comparison between \acrshort{G-REP}, \gls{BBVI}, and \gls{ADVI} in terms of performance on the test set. \acrshort{G-REP} outperforms \gls{BBVI} because the latter has not converged in the allowed time, and it also outperforms \gls{ADVI} because of the variational family it uses.}
  \vspace*{-4pt}
\end{figure}

However, \acrshort{G-REP} provides higher likelihood values than \gls{ADVI}. We show in Figure~\ref{fig:pred_nips} the evolution of the perplexity (lower is better) for the \textsc{nips} dataset, and in Figure~\ref{tab:pred_llh} the resulting test log-likelihood (larger is better) for the rest of the considered datasets. In Figure~\ref{tab:pred_llh}, we report the mean and standard deviation over $100$ posterior samples. \gls{ADVI} cannot fit the data as well as \acrshort{G-REP} or \gls{BBVI} because it is constrained to log-normal and logit-normal variational distributions. These cannot capture sparsity, which is an important feature for the considered models. We can also conclude this by a simple visual inspection of the fitted models. In the Supplement, we compare images sampled from the \acrshort{G-REP} and the \gls{ADVI} posteriors, where we can observe that the latter are more blurry or lack some details.

\vspace*{-9pt}
\section{Conclusion}
\vspace*{-9pt}

We have introduced the generalized reparameterization gradient (\acrshort{G-REP}), a technique to extend the standard reparameterization gradient to a wider class of variational distributions. As the standard reparameterization method, our method is applicable to any probabilistic model that is differentiable with respect to the latent variables. We have demonstrated the generalized reparameterization gradient on two nonconjugate probabilistic models to fit a variational approximation involving gamma and beta distributions. We have also empirically shown that a single Monte Carlo sample is enough to obtain a noisy estimate of the gradient, therefore leading to a fast inference procedure.


\vspace*{-7pt}
\subsubsection*{Acknowledgments}
\vspace*{-5pt}
This project has received funding from the EU H2020 programme (Marie Sk\l{}odowska-Curie grant agreement 706760), NFS IIS-1247664, ONR N00014-11-1-0651, DARPA FA8750-14-2-0009, DARPA N66001-15-C-4032, Adobe, the John Templeton Foundation, and the Sloan Foundation. The authors would also like to thank Kriste Krstovski, Alp Kuckukelbir, and Christian Naesseth for helpful comments and discussions.

\vspace*{-10pt}
\small
\bibliographystyle{apa}
\bibliography{bibReparamGrad}


\newpage
\appendix

\section{Derivation of the Generalized Reparameterization Gradient}

Here we show the mathematical derivation of the generalized reparameterization gradient. Firstly, recall the definition of the functions 
\begin{align}
    & \hfun(\beps;\bv)\triangleq \nabla_{\bv} \Tcal(\beps;\bv), \\
    & \ufun(\beps;\bv)\triangleq \nabla_{\bv} \log J(\beps,\bv),
\end{align}
which are provided in the main text.

We start from the following expression of the gradient, also derived in the main text:
\begin{equation}
    \nabla_{\bv} \E{q(\bz;\bv)}{f(\bz)} = \underbrace{\int q_{\beps}(\beps;\bv)\nabla_{\bv}f\left(\Tcal(\beps;\bv)\right)d\beps}_{\bg^{\textrm{rep}}} +
    \underbrace{\int q_{\beps}(\beps;\bv)f\left(\Tcal(\beps;\bv)\right)\nabla_{\bv} \log q_{\beps}(\beps;\bv) d\beps}_{\bg^{\textrm{corr}}},
\end{equation}

We can write the former term, $\bg^{\textrm{rep}}$, as
\begin{align}
    \bg^{\textrm{rep}} & = \int q_{\beps}(\beps;\bv)\nabla_{\bv}f\left(\Tcal(\beps;\bv)\right)d\beps \\ 
    & = \int q\left(\Tcal(\beps;\bv);\bv\right)J(\beps,\bv) \nabla_{\bv}f\left(\Tcal(\beps;\bv)\right)d\beps \\
    & = \int q\left(\Tcal(\beps;\bv);\bv\right)J(\beps,\bv) \nabla_{\bz}f(\bz)\big|_{\bz=\Tcal(\beps;\bv)} \nabla_{\bv} \Tcal(\beps;\bv)d\beps \\
    & = \int q(\bz;\bv) \nabla_{\bz}f(\bz) \hfun\left( \Tcal^{-1}(\bz;\bv);\bv \right) d\bz \\
    & = \E{q(\bz;\bv)}{\nabla_{\bz}f(\bz) \hfun\left( \Tcal^{-1}(\bz;\bv);\bv \right)},
\end{align}
where we have first replaced the variational distribution on the transformed space with its form as a function of $q(\bz;\bv)$, i.e., $q_{\beps}(\beps;\bv) = q\left(\Tcal(\beps;\bv);\bv\right)J(\beps,\bv)$. We have then applied the chain rule, and finally we have made a new change of variables back to the original space $\bz$ (thus multiplying by the inverse Jacobian).

For the latter, $\bg^{\textrm{corr}}$, we have that
\begin{align}
    \bg^{\textrm{corr}} & = \int q_{\beps}(\beps;\bv)f\left(\Tcal(\beps;\bv)\right)\nabla_{\bv} \log q_{\beps}(\beps;\bv) d\beps \\ 
    & = \int q\left(\Tcal(\beps,\bv);\bv\right)J(\beps,\bv)f\left(\Tcal(\beps;\bv)\right)\nabla_{\bv} \left(\log q\left(\Tcal(\beps;\bv);\bv\right)+\log J(\beps,\bv)\right) d\beps \\ 
    & = \int q\left(\Tcal(\beps;\bv);\bv\right)J(\beps,\bv)f\left(\Tcal(\beps;\bv)\right)\left( \nabla_{\bv} \log q\left(\Tcal(\beps;\bv);\bv\right)+\nabla_{\bv} \log J(\beps,\bv)\right) d\beps. 
\end{align}
The derivative $\nabla_{\bv}\log q\left(\Tcal(\beps;\bv);\bv\right)$ can be obtained by the chain rule. If $\bz=\Tcal(\beps;\bv)$, then $\nabla_{\bv}\log q\left(\Tcal(\beps;\bv);\bv\right)=\nabla_{\bz}\log q(\bz;\bv) \nabla_{\bv}\Tcal(\beps;\bv) + \nabla_{\bv}\log q(\bz;\bv)$. We substitute this result in the above equation and revert the change of variables back to the original space $\bz$ (also multiplying by the inverse Jacobian), yielding
\begin{equation}\label{eq:g_corr}
  \begin{split}
    \bg^{\textrm{corr}} & = \int q(\bz;\bv) f(\bz) \left(\nabla_{\bz}\log q(\bz;\bv) \hfun\left(\Tcal^{-1}(\bz;\bv);\bv\right) + \nabla_{\bv}\log q(\bz;\bv)+ \ufun\left(\Tcal^{-1}(\bz;\bv);\bv \right) \right) d\bz \\
    & = \E{q(\bz;\bv)}{f(\bz) \left(\nabla_{\bz}\log q(\bz;\bv) \hfun\left(\Tcal^{-1}(\bz;\bv);\bv\right) + \nabla_{\bv}\log q(\bz;\bv)+ \ufun\left(\Tcal^{-1}(\bz;\bv);\bv \right) \right)},
  \end{split}
\end{equation}
where we have used the definition of the functions $\hfun(\beps;\bv)$ and $\ufun(\beps;\bv)$.

\section{Particularization for the Gamma Distribution}

For the gamma distribution we choose the transformation
\begin{align}
z=\Tcal(\epsilon;\alpha,\beta)=\exp(\epsilon\sqrt{\psi_1(\alpha)}+\psi(\alpha)-\log(\beta)).
\end{align}
Thus, we have that
\begin{align}
    & J(\epsilon,\alpha,\beta) = |\det \nabla_{\epsilon}\Tcal(\epsilon;\alpha,\beta)| = \Tcal(\epsilon;\alpha,\beta)\sqrt{\psi_1(\alpha)}.
\end{align}
The derivatives of $\log q(z;\alpha,\beta)$ with respect to its arguments are given by
\begin{align}
    & \frac{\partial}{\partial z}\log q(z;\alpha,\beta) = \frac{\alpha-1}{z}-\beta,\\
    & \frac{\partial}{\partial \alpha}\log q(z;\alpha,\beta) = \log(\beta)-\psi(\alpha)+\log(z),\\
    & \frac{\partial}{\partial \beta}\log q(z;\alpha,\beta) = \frac{\alpha}{\beta}-z.
\end{align}

Therefore, the auxiliary functions $\hfun(\epsilon;\alpha,\beta)$ and $\ufun(\epsilon;\alpha,\beta)$ for the components of the gradient with respect to $\alpha$ and $\beta$ can be written as
\begin{align}
    & \hfun_{\alpha}(\epsilon;\alpha,\beta) = \frac{\partial}{\partial\alpha}\Tcal(\epsilon;\alpha,\beta) = \Tcal(\epsilon;\alpha,\beta) \left( \frac{\epsilon\psi_2(\alpha)}{2\sqrt{\psi_1(\alpha)}}+\psi_1(\alpha)\right),\\
    & \hfun_{\beta}(\epsilon;\alpha,\beta) = \frac{\partial}{\partial\beta}\Tcal(\epsilon;\alpha,\beta) = -\frac{\Tcal(\epsilon;\alpha,\beta)}{\beta},\\
    & \ufun_{\alpha}(\epsilon;\alpha,\beta) = \frac{\partial}{\partial\alpha}\log J(\epsilon,\alpha,\beta) = \left( \frac{\epsilon\psi_2(\alpha)}{2\sqrt{\psi_1(\alpha)}}+\psi_1(\alpha)\right)+\frac{\psi_2(\alpha)}{2\psi_1(\alpha)},\\
    & \ufun_{\beta}(\epsilon;\alpha,\beta) = \frac{\partial}{\partial\beta}\log J(\epsilon,\alpha,\beta) = -\frac{1}{\beta}.
\end{align}

Thus, we finally obtain that the components of $\bg^{\textrm{rep}}$ corresponding to the derivatives with respect to $\alpha$ and $\beta$ are given by
\begin{align}
    & \bg^{\textrm{rep}}_{\alpha} = \E{q(z;\alpha,\beta)}{\frac{\partial}{\partial z}f(\bz) \times z \left( \frac{\Tcal^{-1}(z;\alpha,\beta)\psi_2(\alpha)}{2\sqrt{\psi_1(\alpha)}}+\psi_1(\alpha)\right)},\\
    & \bg^{\textrm{rep}}_{\beta} = \E{q(z;\alpha,\beta)}{\frac{\partial}{\partial z}f(\bz) \times \frac{-z}{\beta}},
\end{align}
while the components of $\bg^{\textrm{corr}}$ can be similarly obtained by substituting the expressions above into Eq.~\ref{eq:g_corr}. Remarkably, we obtain that 
\begin{align}
    & \bg^{\textrm{corr}}_{\beta} = 0.
\end{align}

\section{Particularization for the Beta Distribution}

For a random variable $z\sim\textrm{Beta}(\alpha,\beta)$, we could rewrite $z=z_1^\prime/(z_1^\prime+z_2^\prime)$ for $z_1^\prime\sim\textrm{Gamma}(\alpha,1)$ and $z_2^\prime\sim\textrm{Gamma}(\beta,1)$, and apply the above method for the gamma-distributed variables $z_1^\prime$ and $z_2^\prime$. Instead, in the spirit of applying standardization directly over $z$, we define a transformation to standardize the logit function. This leads to
\begin{align}
    z=\Tcal(\epsilon;\alpha,\beta)=\frac{1}{1+\exp(-\epsilon\sigma-\psi(\alpha)+\psi(\beta))}.
\end{align}
This transformation ensures that $\epsilon$ has mean zero. However, in this case we do not specify the form of $\sigma$, and we let it be a function of $\alpha$ and $\beta$. This allows us to choose $\sigma$ in such a way that $\bg^{\textrm{corr}}=\bzero$ for the sampled value of $z$, which we found to work well (even though this introduces some bias). For simplicity, we write $\sigma=\exp(\phi)$.

Thus, we have that
\begin{align}
    & J(\epsilon,\alpha,\beta) = |\det \nabla_{\epsilon}\Tcal(\epsilon;\alpha,\beta)| = \Tcal(\epsilon;\alpha,\beta)(1-\Tcal(\epsilon;\alpha,\beta))\sigma.
\end{align}
The derivatives of $\log q(z;\alpha,\beta)$ with respect to its arguments are given by
\begin{align}
    & \frac{\partial}{\partial z}\log q(z;\alpha,\beta) = \frac{\alpha-1}{z} - \frac{\beta-1}{1-z},\\
    & \frac{\partial}{\partial \alpha}\log q(z;\alpha,\beta) = \psi(\alpha+\beta)-\psi(\alpha)+\log(z),\\
    & \frac{\partial}{\partial \beta}\log q(z;\alpha,\beta) = \psi(\alpha+\beta)-\psi(\beta)+\log(1-z).
\end{align}

Therefore, the auxiliary functions $\hfun(\epsilon;\alpha,\beta)$ and $\ufun(\epsilon;\alpha,\beta)$ for the components of the gradient with respect to $\alpha$ and $\beta$ can be written as
\begin{align}
    & \hfun_{\alpha}(\epsilon;\alpha,\beta) = \frac{\partial}{\partial\alpha}\Tcal(\epsilon;\alpha,\beta) = \Tcal(\epsilon;\alpha,\beta)(1-\Tcal(\epsilon;\alpha,\beta))\left( \psi_1(\alpha) + \epsilon\sigma\frac{\partial\phi}{\partial\alpha} \right) ,\\
    & \hfun_{\beta}(\epsilon;\alpha,\beta) = \frac{\partial}{\partial\beta}\Tcal(\epsilon;\alpha,\beta) = \Tcal(\epsilon;\alpha,\beta)(1-\Tcal(\epsilon;\alpha,\beta))\left( -\psi_1(\beta) + \epsilon\sigma\frac{\partial\phi}{\partial\beta} \right),\\
    & \ufun_{\alpha}(\epsilon;\alpha,\beta) = \frac{\partial}{\partial\alpha}\log J(\epsilon,\alpha,\beta) = (1-2\Tcal(\epsilon;\alpha,\beta))\left( \psi_1(\alpha) + \epsilon\sigma\frac{\partial\phi}{\partial\alpha} \right)+\frac{\partial\phi}{\partial\alpha},\\
    & \ufun_{\beta}(\epsilon;\alpha,\beta) = \frac{\partial}{\partial\beta}\log J(\epsilon,\alpha,\beta) = (1-2\Tcal(\epsilon;\alpha,\beta))\left( -\psi_1(\beta) + \epsilon\sigma\frac{\partial\phi}{\partial\beta} \right)+\frac{\partial\phi}{\partial\beta}.
\end{align}
Note that the term $\epsilon\sigma$ above can be computed from $z$ without knowledge of the value of $\sigma$ as $\epsilon\sigma=\Tcal^{-1}(z;\alpha,\beta)\sigma = \frac{\logit{z}-\psi(\alpha)+\psi(\beta)}{\sigma}\sigma=\logit{z}-\psi(\alpha)+\psi(\beta)$.

Thus, we finally obtain that the components of $\bg^{\textrm{rep}}$ corresponding to the derivatives with respect to $\alpha$ and $\beta$ are given by
\begin{align}
    & \bg^{\textrm{rep}}_{\alpha} = \E{q(z;\alpha,\beta)}{\frac{\partial}{\partial z}f(\bz) \times z(1-z)\left( \psi_1(\alpha) + \left(\logit{z}-\psi(\alpha)+\psi(\beta)\right)\frac{\partial\phi}{\partial\alpha} \right)},\\
    & \bg^{\textrm{rep}}_{\beta} = \E{q(z;\alpha,\beta)}{\frac{\partial}{\partial z}f(\bz) \times z(1-z)\left( -\psi_1(\beta) + \left(\logit{z}-\psi(\alpha)+\psi(\beta)\right)\frac{\partial\phi}{\partial\beta} \right)},
\end{align}
where we are still free to choose $\partial\phi/\partial\alpha$ and $\partial\phi/\partial\beta$. We have found that the choice of these values such that $\bg^{\textrm{corr}}_{\alpha} = \bg^{\textrm{corr}}_{\beta} = 0$ works well in practice. Thus, we set the derivatives of $\phi$ such that the relationships
\begin{align}
    \frac{\partial}{\partial z}\log q(z;\alpha,\beta) \times \hfun_{\alpha}\left(\Tcal^{-1}(z;\alpha,\beta);\alpha,\beta \right) + \frac{\partial}{\partial \alpha}\log q(z;\alpha,\beta) + \ufun_{\alpha}\left(\Tcal^{-1}(z;\alpha,\beta);\alpha,\beta \right) = 0,\\
    \frac{\partial}{\partial z}\log q(z;\alpha,\beta) \times \hfun_{\beta}\left(\Tcal^{-1}(z;\alpha,\beta);\alpha,\beta \right) + \frac{\partial}{\partial \beta}\log q(z;\alpha,\beta) + \ufun_{\beta}\left(\Tcal^{-1}(z;\alpha,\beta);\alpha,\beta \right) = 0,
\end{align}
hold for the sampled value of $z$. This involves solving a simple linear equation for $\partial\phi/\partial\alpha$ and $\partial\phi/\partial\beta$.

\section{Particularization for the Dirichlet Distribution}

For a $\textrm{Dirichlet}(\balpha)$ distribution, with $\balpha=[\alpha_1,\ldots,\alpha_K]$, we can apply the standardization
\begin{equation}
\bz=\Tcal(\beps;\balpha) = \exp\left(\bSigma^{1/2}\beps+\bmu\right),
\end{equation}
where the mean $\bmu$ is a $K$-length vector and the covariance $\bSigma$ is a $K\times K$ matrix,\footnote{Instead, we could define a transformation that ignores the off-diagonal terms of the covariance matrix. This would lead to faster computations but higher variance of the resulting estimator.}$^{,}$\footnote{We could also apply the full-covariance transformation for the beta distribution.} which are respectively given by
\begin{equation}
 \begin{split}
  \bmu = \E{q(\bz;\balpha)}{\log(\bz)} =
  \left[
  \begin{array}{c}
   \psi(\alpha_1)-\psi(\alpha_0) \\
   \vdots \\
   \psi(\alpha_K)-\psi(\alpha_0) \\
   \end{array}
  \right]
 \end{split}
\end{equation}
and
\begin{equation}
 (\bSigma)_{ij} = \textrm{Cov}(\log(z_i),\log(z_j)) =
 \left\{
  \begin{array}{ll}
   \psi_1(\alpha_i)-\psi_1(\alpha_0) & \textrm{if } i=j, \\
   -\psi_1(\alpha_0) & \textrm{if } i\neq j. \\
  \end{array}
 \right.
\end{equation}
Here, we have defined $\alpha_0=\sum_k \alpha_k$. The covariance matrix $\bSigma$ can be rewritten as a diagonal matrix plus a rank one update, which can be exploited for faster computations:
\begin{equation}
 \bSigma = \textrm{diag}\left(
  \left[
  \begin{array}{c}
   \psi_1(\alpha_1) \\
   \vdots \\
   \psi_1(\alpha_K) \\
   \end{array}
  \right]
 \right) - \psi_1(\alpha_0) \bone\bone^\top.
\end{equation}
Note that, since $\bSigma$ is positive semidefinite, $\bSigma^{1/2}$ can be readily obtained after diagonalization. In other words, if we express $\bSigma = \bV \bD \bV^\top$, where $\bV$ is an orthonormal matrix and $\bD$ is a diagonal matrix, then $\bSigma^{1/2} = \bV \bD^{1/2} \bV^\top$.

Given the transformation above, we can write
\begin{equation}
 J(\beps,\balpha) = |\det \nabla_{\beps}\Tcal(\beps;\balpha)| = \det(\bSigma^{1/2}) \prod_i \Tcal_i(\beps;\balpha).
\end{equation}

The derivatives of $\log q(\bz;\balpha)$ with respect to its arguments are given by
\begin{align}
    & \frac{\partial}{\partial z_i}\log q(\bz;\balpha) = \frac{\alpha_i-1}{z_i},\\
    & \frac{\partial}{\partial \alpha_i}\log q(\bz;\balpha) = \psi(\alpha_0)-\psi(\alpha_i)+\log(z_i).
\end{align}

Therefore, the auxiliary functions $\hfun(\beps;\balpha)$ and $\ufun(\beps;\balpha)$ can be written as
\begin{align}
    & \hfun(\beps;\balpha) = \nabla_{\balpha}\Tcal(\beps;\balpha) = 
    \left[
     \begin{array}{ccc}
       \Tcal_1(\beps;\balpha)\left(\frac{\partial(\bSigma_{1:}^{1/2})}{\partial\alpha_1}\beps+\frac{\partial\mu_1}{\partial\alpha_1}\right) & \cdots & \Tcal_1(\beps;\balpha)\left(\frac{\partial(\bSigma_{1:}^{1/2})}{\partial\alpha_K}\beps+\frac{\partial\mu_1}{\partial\alpha_K}\right) \\
       \vdots & \ddots & \vdots \\
       \Tcal_{K}(\beps;\balpha)\left(\frac{\partial(\bSigma_{K:}^{1/2})}{\partial\alpha_1}\beps+\frac{\partial\mu_{K}}{\partial\alpha_1}\right) & \cdots & \Tcal_{K}(\beps;\balpha)\left(\frac{\partial(\bSigma_{K:}^{1/2})}{\partial\alpha_K}\beps+\frac{\partial\mu_{K}}{\partial\alpha_K}\right) \\
     \end{array}
    \right]
     ,\\
    & \ufun(\beps;\balpha) = \nabla_{\balpha}\log J(\beps,\balpha) = 
    \left[
     \begin{array}{c}
       \frac{\partial \log\det(\bSigma^{1/2})}{\partial\alpha_1} + \sum_i \left(\frac{\partial(\bSigma_{i:}^{1/2})}{\partial\alpha_1}\beps+\frac{\partial\mu_i}{\partial\alpha_1}\right)
        \\
       \vdots \\
       \frac{\partial \log\det(\bSigma^{1/2})}{\partial\alpha_K} + \sum_i \left(\frac{\partial(\bSigma_{i:}^{1/2})}{\partial\alpha_K}\beps+\frac{\partial\mu_i}{\partial\alpha_K}\right)
     \end{array}
    \right].
\end{align}

The intermediate derivatives that are necessary for the computation of the functions $\hfun(\beps;\balpha)$ and $\ufun(\beps;\balpha)$ are:
\begin{align}
    & \frac{\partial\bmu}{\partial\alpha_i} = 
  \left[
  \begin{array}{c}
   -\psi_1(\alpha_0) \\
   -\psi_1(\alpha_0) \\
   \vdots \\
   \psi_1(\alpha_i)-\psi_1(\alpha_0) \\
   \vdots \\
   -\psi_1(\alpha_0) \\
   \end{array}
  \right] \\
    & \frac{\partial \log\det(\bSigma^{1/2})}{\partial\alpha_i} = \textrm{trace}\left( \bSigma^{-1/2} \frac{\partial \bSigma^{1/2}}{\partial\alpha_i} \right),
\end{align}
and $\frac{\partial \bSigma^{1/2}}{\partial\alpha_i}$ is the solution to the Lyapunov equation
\begin{equation}
 \frac{\partial \bSigma}{\partial\alpha_i} = \frac{\partial \bSigma^{1/2}}{\partial\alpha_i} \bSigma^{1/2}+\bSigma^{1/2}\frac{\partial \bSigma^{1/2}}{\partial\alpha_i},
\end{equation}
where
\begin{equation}
 \frac{\partial\bSigma}{\partial\alpha_i} = \textrm{diag}\left(
  \left[
  \begin{array}{c}
   0 \\
   0 \\
   \vdots \\
   \psi_2(\alpha_i) \\
   \vdots \\
   0 \\
   \end{array}
  \right]
 \right) - \psi_2(\alpha_0) \bone\bone^\top.
\end{equation}

Putting all this together, we finally have the expressions for the generalized reparameterization gradient:
\begin{align}
    \bg^{\textrm{rep}} & = \E{q(\bz;\balpha)}{\hfun^\top\left( \Tcal^{-1}(\bz;\balpha);\balpha \right)\nabla_{\bz}f(\bz)},\\
    \bg^{\textrm{corr}} & = \E{q(\bz;\balpha)}{f(\bz) \Big( \hfun^\top\left(\Tcal^{-1}(\bz;\balpha);\balpha\right)\nabla_{\bz}\log q(\bz;\bv) + \nabla_{\balpha}\log q(\bz;\balpha)+ \ufun\left(\Tcal^{-1}(\bz;\balpha);\balpha \right) \Big)},
\end{align}

\section{Experimental Results}

\subsection{Using more than 1 sample}

\begin{figure}[ht]
  \centering
  \subfloat[\gls{ELBO}.]{\includegraphics[width=0.36\textwidth]{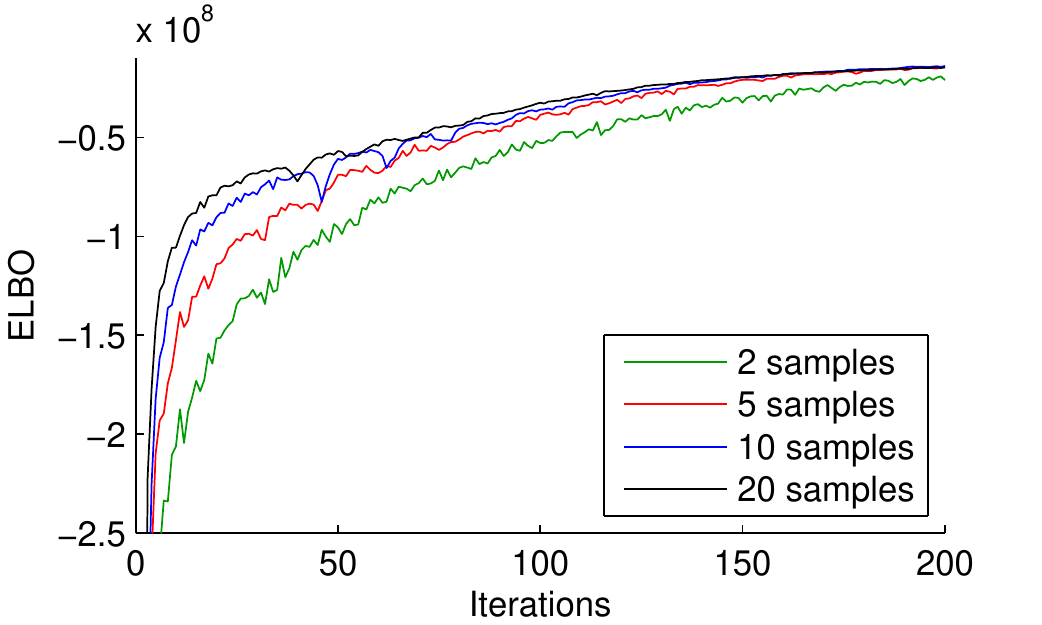}} \qquad
  \subfloat[Average variance.]{\includegraphics[width=0.36\textwidth]{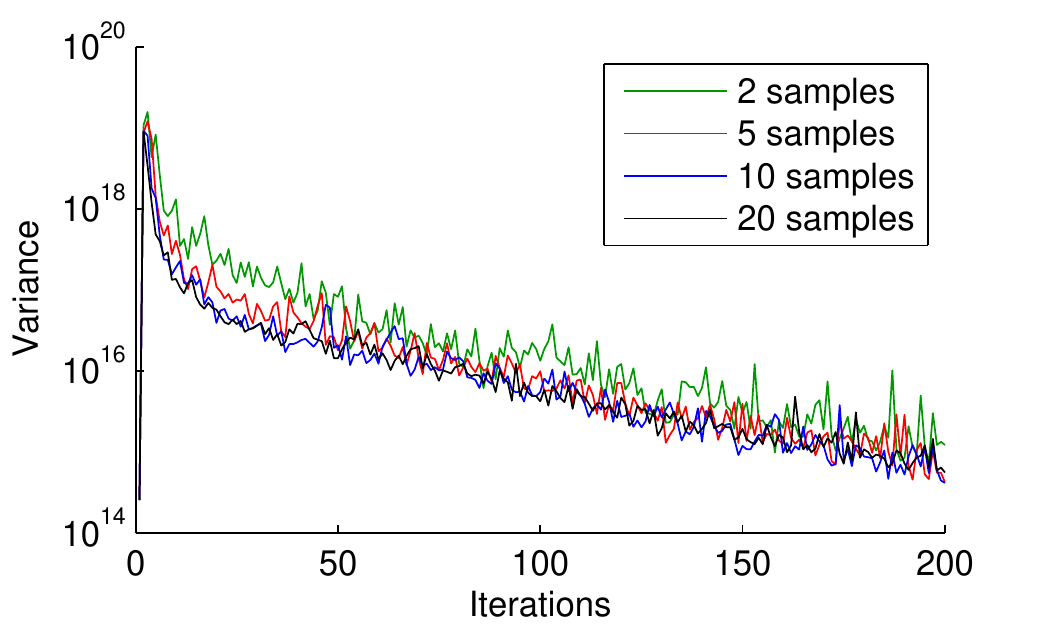}}
  \caption{Performance of \acrshort{G-REP} for different number of Monte Carlo samples.\label{fig:nsamples}}
\end{figure}

We now study the sensitivity of the generalized reparameterization gradient with respect to the number of samples of the Monte Carlo estimator.
For that, we choose the Olivetti dataset, and we apply the generalized reparameterization approach using 2, 5, 10, and 20 Monte Carlo samples. At each iteration, we compute the \gls{ELBO} and the average sample variance of the gradient estimator. We report these results in Figure~\ref{fig:nsamples} for the first 200 iterations of the inference procedure. As expected, increasing the number of samples is beneficial because it reduces the resulting variance. The gap between the curves with $10$ and $20$ samples is negligible, specially after $100$ iterations. A larger number of samples seems to be particularly helpful in the very early iterations of inference.

\subsection{Reconstructed images}

Here, we show some reconstructed observations for the three datasets involving images, namely, the binarized \textsc{mnist}, the Olivetti dataset, and Omniglot. We plot the reconstructed images as follows: we first draw one sample from the variational posterior, and then we compute the \emph{mean} of the observations for that particular sample of latent variables.

\begin{figure}[ht]
  \centering
  \subfloat[True observations.]{\includegraphics[width=0.33\textwidth]{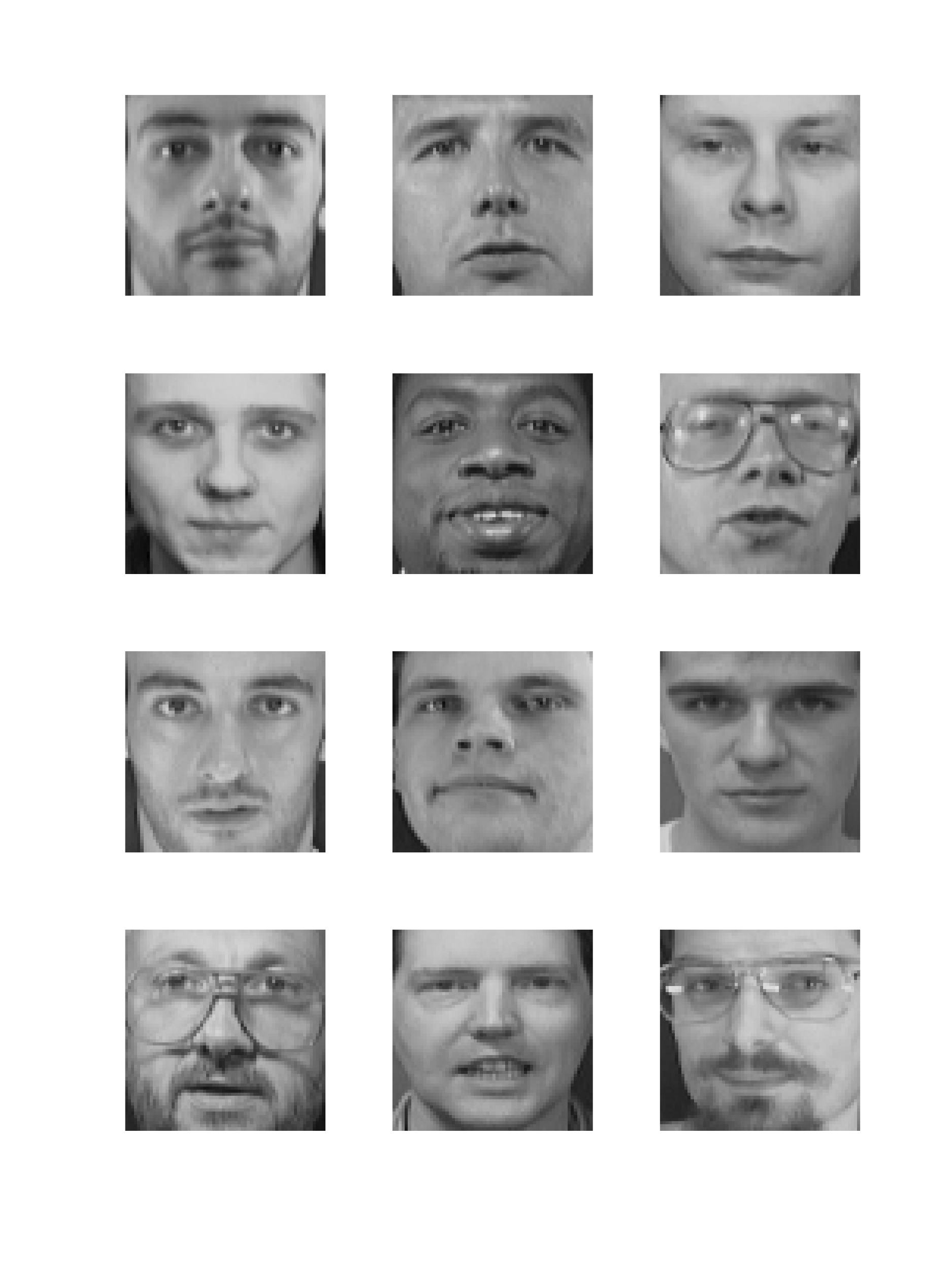}}
  \subfloat[Reconstructed (\acrshort{G-REP}).]{\includegraphics[width=0.33\textwidth]{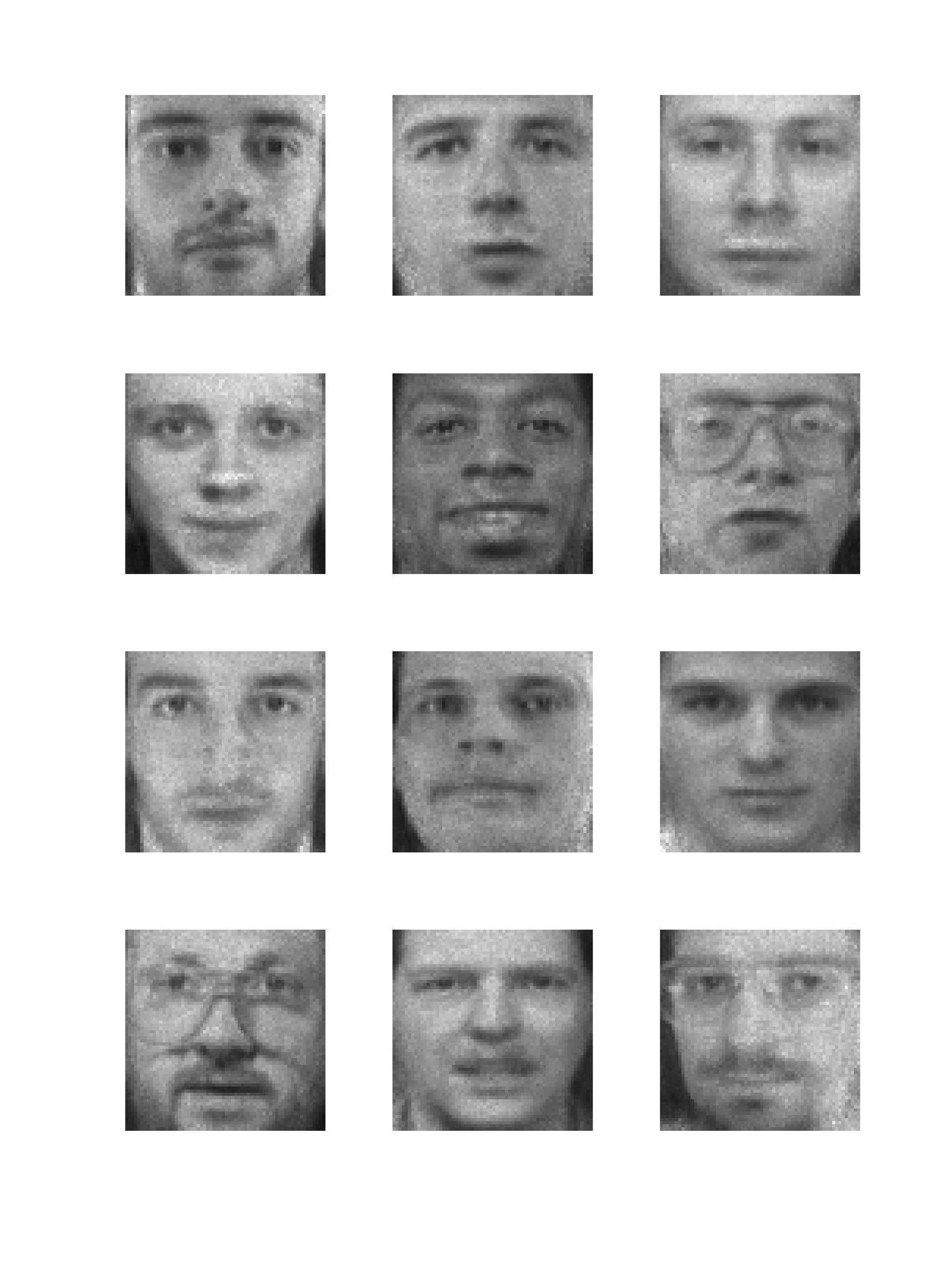}}
  \subfloat[Reconstructed (\acrshort{ADVI}).]{\includegraphics[width=0.33\textwidth]{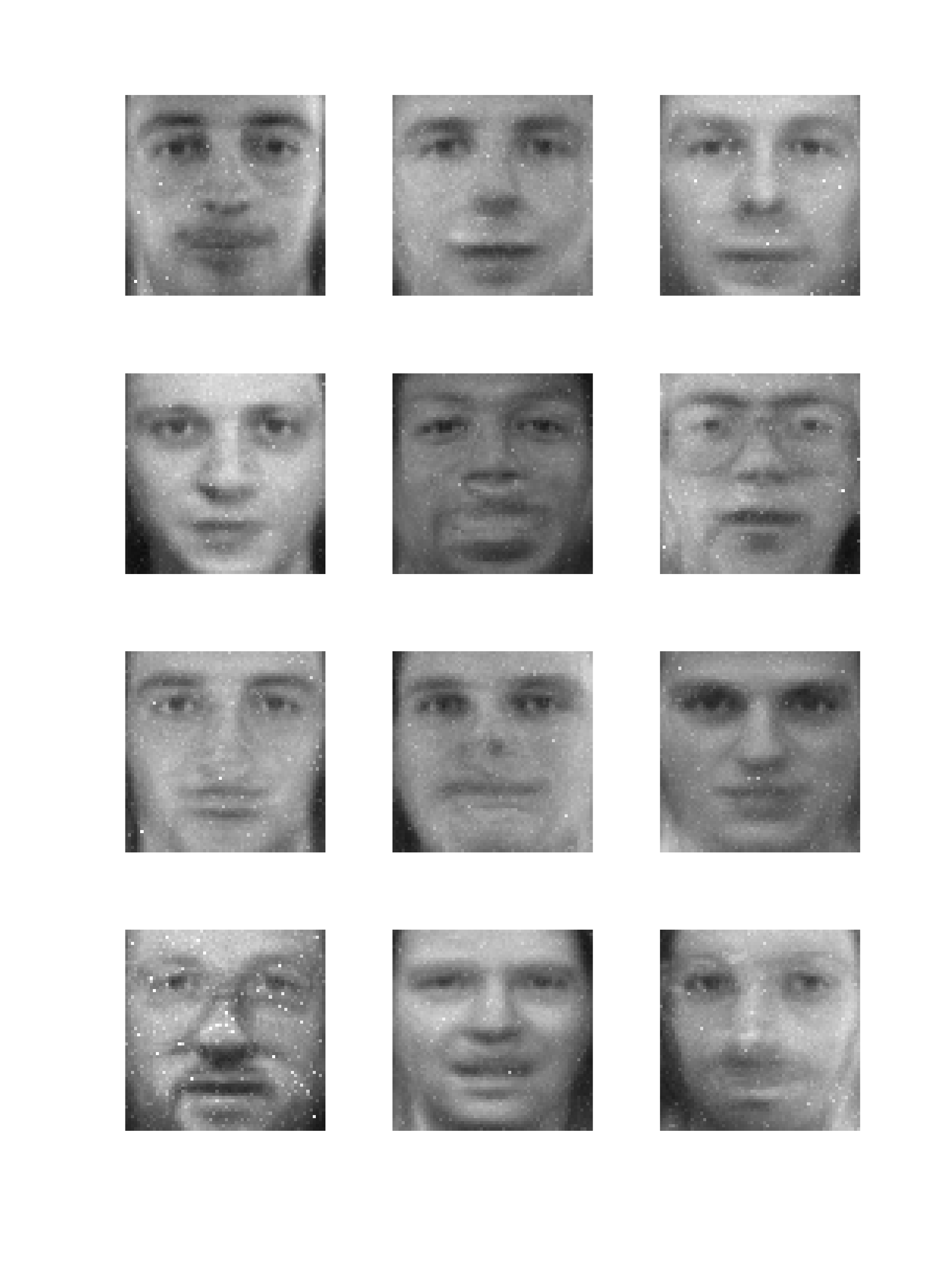}}
  \caption{Images from the Olivetti dataset. \gls{ADVI} provides less detailed images when compared to \acrshort{G-REP}.\label{fig:images_faces}}
\end{figure}

Figure~\ref{fig:images_faces} shows the results for the Olivetti dataset. The true observations are shown in the left panel, whereas the corresponding reconstructed images are shown in the center panel (for \acrshort{G-REP}) and the right panel (for \acrshort{ADVI}). We can observe that the images obtained from \acrshort{G-REP} are more detailed (e.g., we can distinguish the glasses, mustache, or facial expressions) than the images obtained from \acrshort{ADVI}. We argue that this effect is due to the variational family used by \gls{ADVI}, which cannot capture well sparse posterior distributions, for which samples close to $0$ are common.

\begin{figure}[ht]
  \centering
  \subfloat[True observations.]{\includegraphics[width=0.33\textwidth]{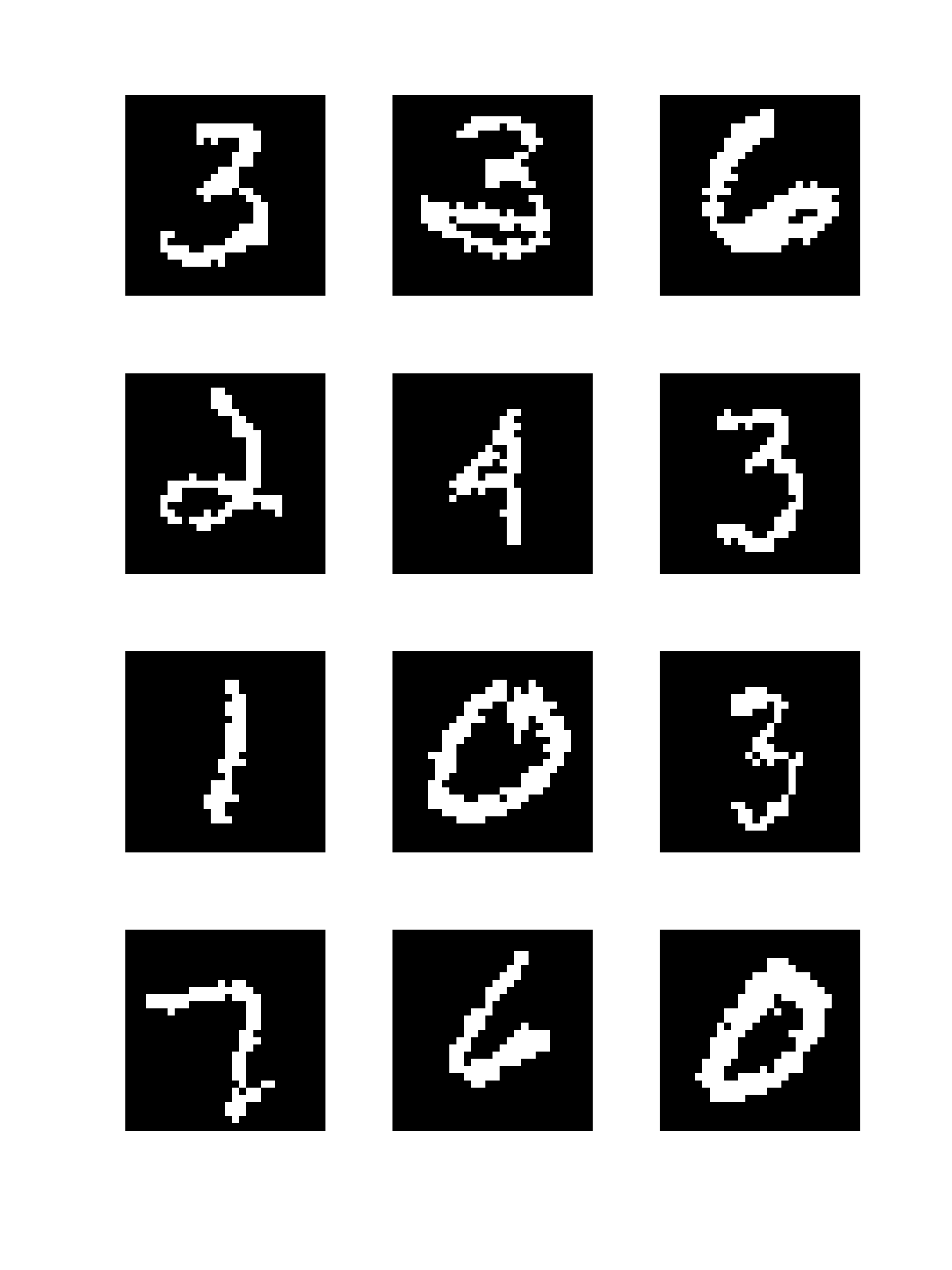}}
  \subfloat[Reconstructed (\acrshort{G-REP}).]{\includegraphics[width=0.33\textwidth]{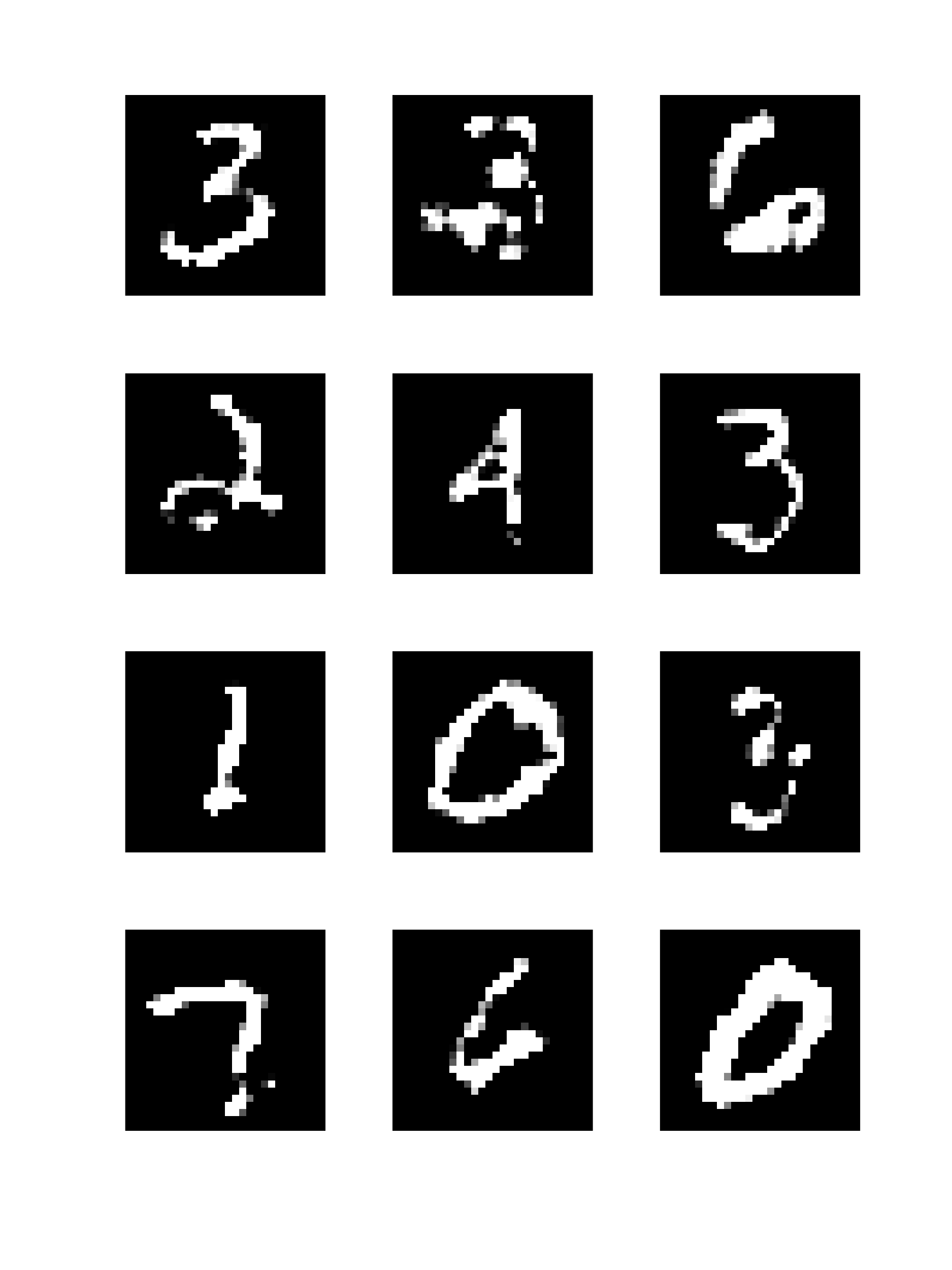}}
  \subfloat[Reconstructed (\acrshort{ADVI}).]{\includegraphics[width=0.33\textwidth]{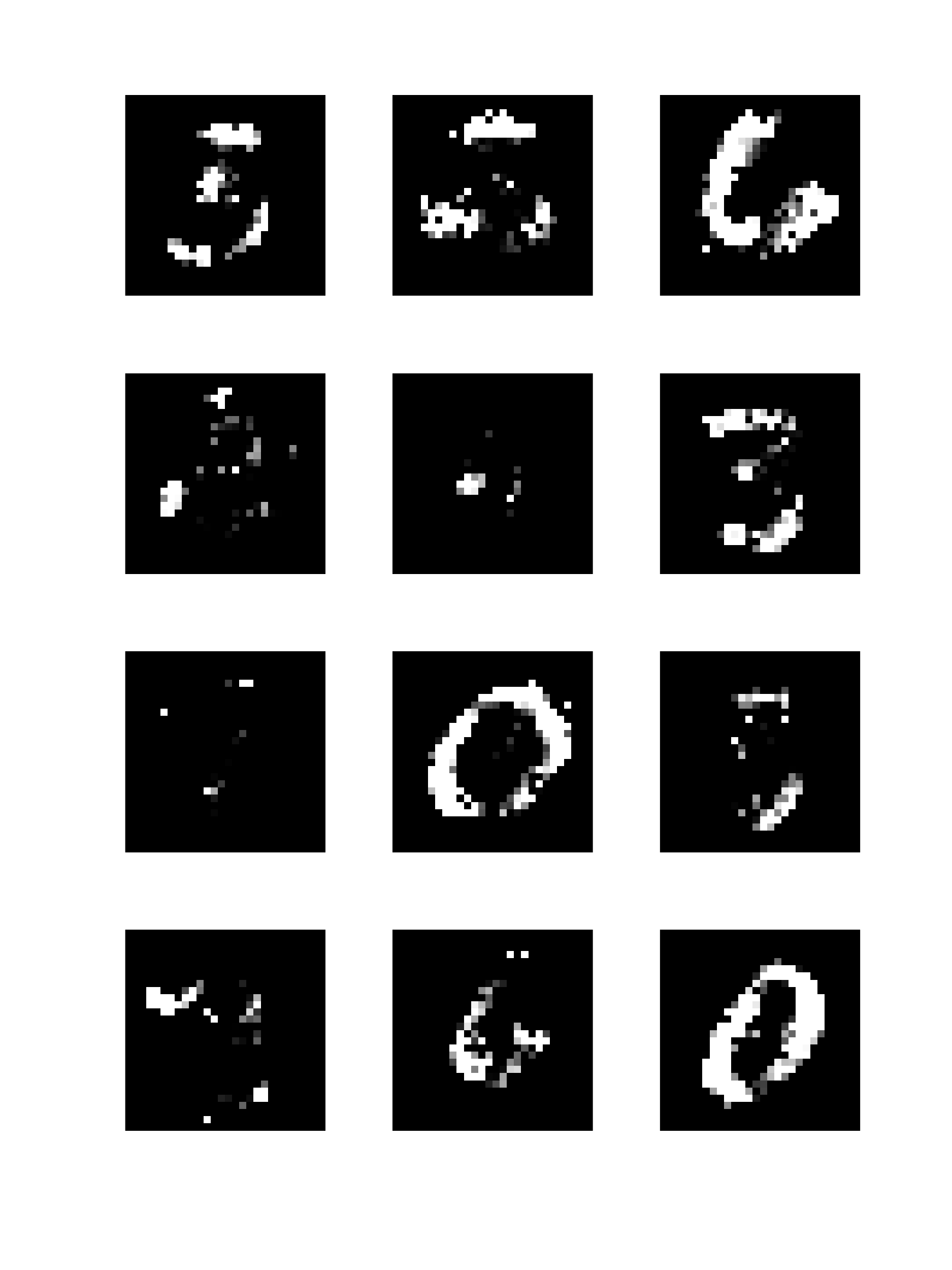}}
  \caption{Images from the binarized \textsc{mnist} dataset. \gls{ADVI} provides more blurry images when compared to \acrshort{G-REP}.\label{fig:images_bmnist}}
\end{figure}

\begin{figure}[ht]
  \centering
  \subfloat[True observations.]{\includegraphics[width=0.33\textwidth]{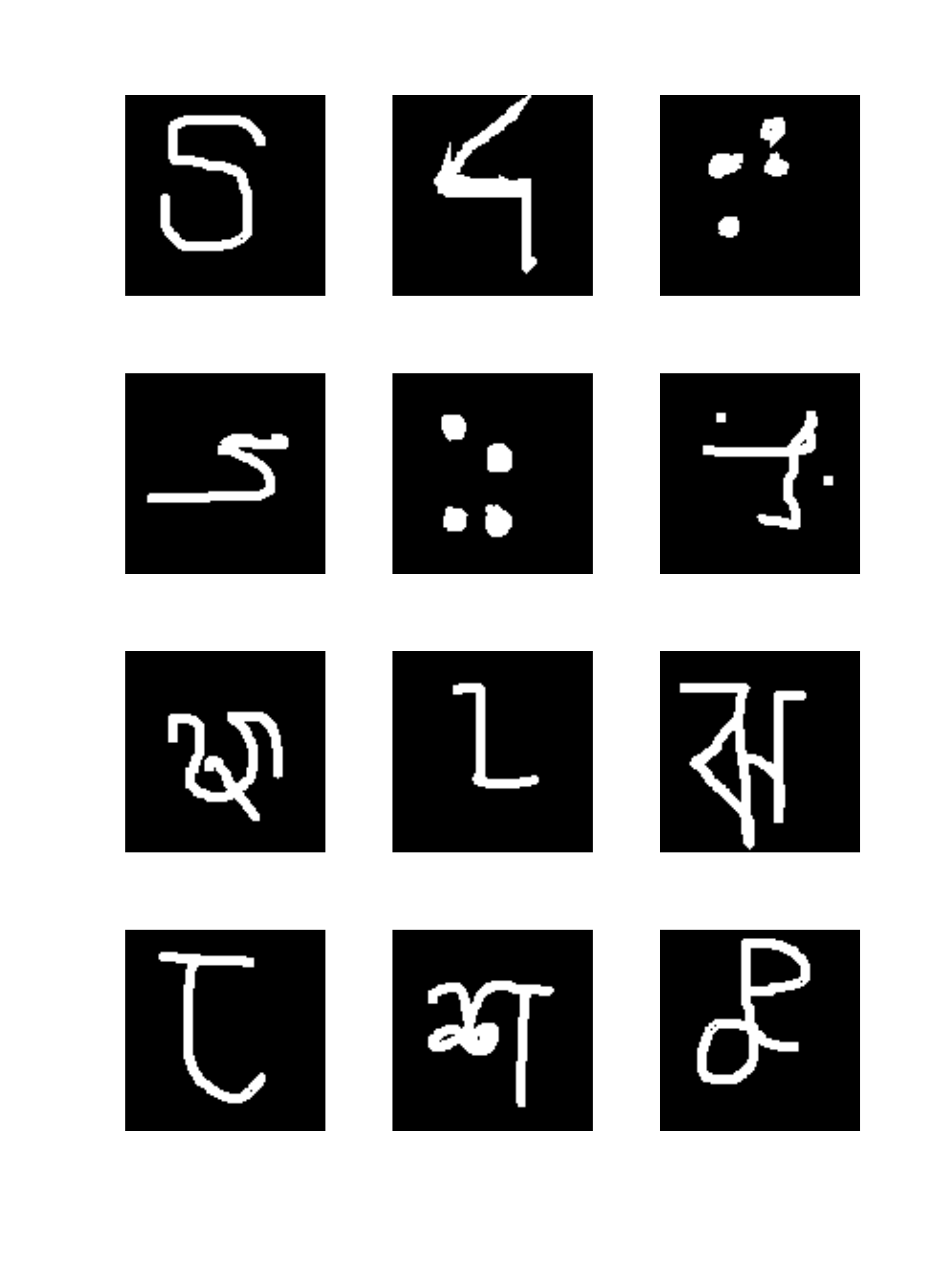}}
  \subfloat[Reconstructed (\acrshort{G-REP}).]{\includegraphics[width=0.33\textwidth]{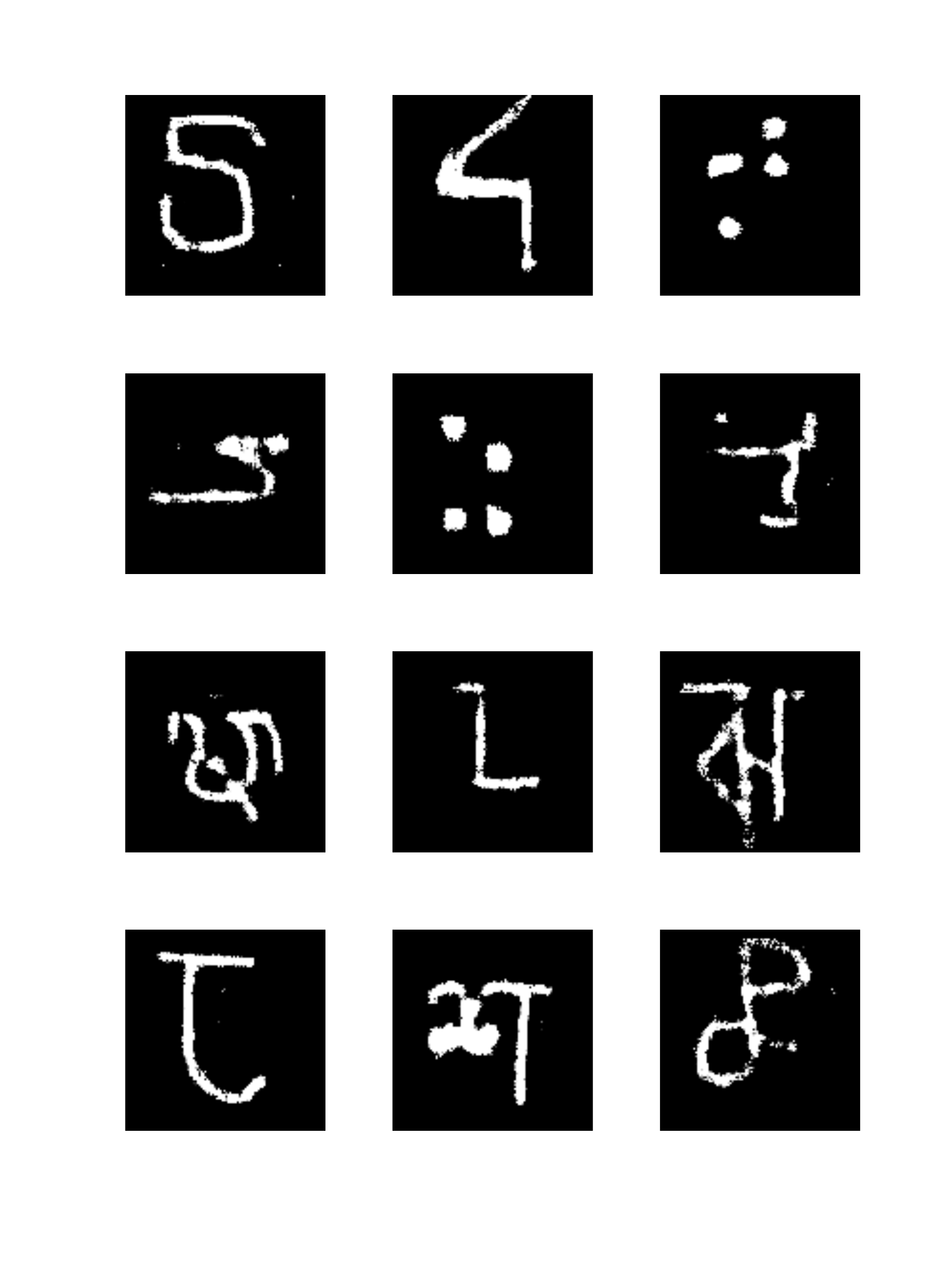}}
  \subfloat[Reconstructed (\acrshort{ADVI}).]{\includegraphics[width=0.33\textwidth]{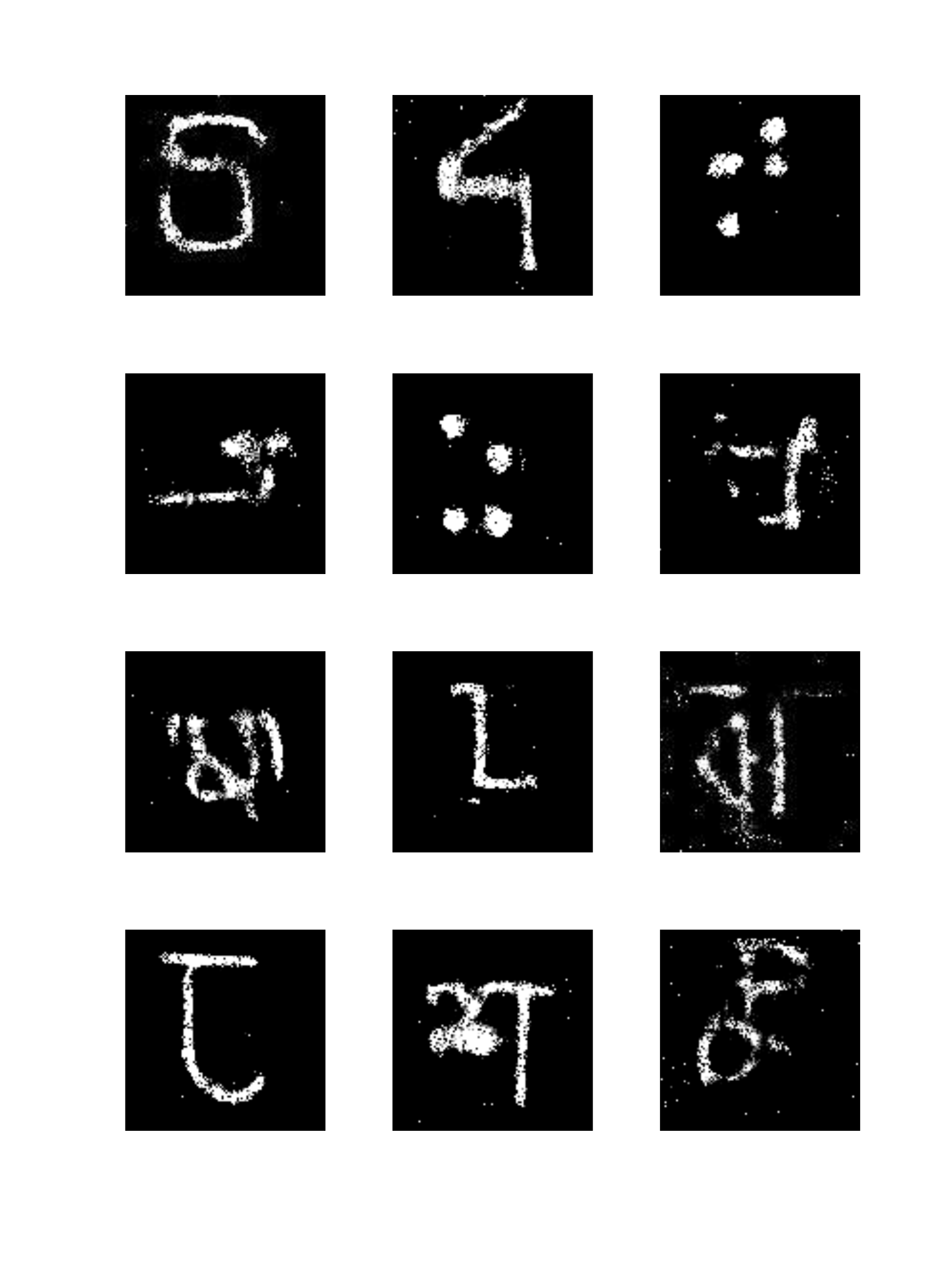}}
  \caption{Images from the Omniglot dataset. \gls{ADVI} provides more blurry images when compared to \acrshort{G-REP}.\label{fig:images_omniglot}}
\end{figure}

This behavior is similar in the case of the digits from \textsc{mnist} or the characters from Omniglot. We show these images in Figures~\ref{fig:images_bmnist} and \ref{fig:images_omniglot}, respectively. Once again, images sampled from the \acrshort{G-REP} posterior are visually closer to the ground truth that images sampled from the \acrshort{ADVI} posterior, which tend to be more blurry, or even unrecognizable in a few cases.

\end{document}